\documentclass{article} %
\usepackage{arxiv}
\usepackage{times}
\usepackage{natbib}

\usepackage[utf8]{inputenc} %
\usepackage[T1]{fontenc}    %
\usepackage{hyperref}       %
\usepackage{url}            %
\usepackage{booktabs}       %
\usepackage{amsfonts}       %
\usepackage{nicefrac}       %
\usepackage{microtype}      %
\usepackage{amsmath}
\usepackage{multirow}
\usepackage{graphicx}
\usepackage{adjustbox}
\usepackage{subfig}
\graphicspath{ {images/} }

\title{Generated Loss, Augmented Training, and Multiscale VAE}

\author{
  Jason Chou\thanks{Work done during the tenure as a Google employee.}, Gautam Hathi \\
  Google LLC. \\
  1600 Amphitheatre Parkway \\
  Mountain View, CA 94043 \\
  \texttt{chuanchih@gmail.com, gautamh@google.com} \\
}

\begin{document}

\maketitle

\begin{abstract}
  The variational autoencoder (VAE) framework remains a popular option for training unsupervised generative models, especially for discrete data where generative adversarial networks (GANs) require workaround to create gradient for the generator. In our work modeling US postal addresses, we show that our discrete VAE with tree recursive architecture demonstrates limited capability of capturing field correlations within structured data, even after overcoming the challenge of posterior collapse with scheduled sampling and tuning of the KL-divergence weight $\beta$. Worse, VAE seems to have difficulty mapping its generated samples to the latent space, as their VAE loss lags behind or even increases during the training process. Motivated by this observation, we show that augmenting training data with generated variants (augmented training) and training a VAE with multiple values of $\beta$ simultaneously (multiscale VAE) both improve the generation quality of VAE. Despite their differences in motivation and emphasis, we show that augmented training and multiscale VAE are actually connected and have similar effects on the model.
\end{abstract}

\section{Introduction}

The variational autoencoder (VAE) framework \citep{kingma2013auto} and generative adversarial network (GAN) framework \citep{goodfellow2014generative} have been the two dominant options for training deep generative models. Despite recent excitement about GAN, VAE remains a popular option, featuring ease of training and wide applicability, with an encoder-decoder pair being the only problem-specific requirement. The VAE encoder encodes training examples into posterior distributions in an abstract latent space, from which sampled latent vectors are drawn and the VAE decoder is trained to reconstruct the training examples from their respective latent vectors. In addition to minimizing mistakes in reconstruction (`reconstruction loss'), VAE features the competing objective of minimizing the difference between the posterior distributions and an assumed prior (`latent loss', measured by KL-divergence). The total VAE loss the model tries to minimize is the sum of these two losses, and the competing objective of minimizing the latent loss creates an information bottleneck between the encoder and the decoder. Ideally, the learned compression allows a random vector in the latent space to be decoded into a realistic sample in generation time.

Compared to GAN, VAE is directly trained to encode all training examples and therefore is less prone to the failure mode of generating a few memorized training examples (`mode collapse'). On the other hand, it tends to have lower precision, which manifests as blurry images for visual problems \citep{sajjadi2018assessing}. It has been theorized that such blurred reconstructions correspond to multiple distinct training examples and are due their overlapping posterior distributions in the latent space. Conversely, holes in the latent space that do not correspond to any posterior distributions of training examples may result in generated samples unconstrained by training data \citep{rezende2018taming}. One may note that these two issues are two sides of the same coin: Strong information bottleneck leads to too much noise in the sampled latent vectors and overlapping posterior distributions, whereas weak information bottleneck leads to too little noise and leaves holes in the latent space. Unsurprisingly, the simplest approach to improving the VAE has been fine-tuning the strength of the information bottleneck by the introduction of the KL-divergence weight $\beta$ as a hyperparameter. In addition to the hyperparameter sweep of KL-divergence weight $\beta$ \citep{higgins2017beta}, manual annealing in both directions: KL-divergence warm-up \citep{DBLP:journals/corr/BowmanVVDJB15,sonderby2016ladder,DBLP:journals/corr/abs-1804-02135} and controlled capacity increase \citep{burgess2018understanding} has been employed to achieve good latent space structure and accurate reconstruction simultaneously. Such training scheme relies on the model's memory of training steps with a different KL-divergence weight, even though there is no \textit{a priori} reason to prefer any particular one in this case. This generalized $\beta$-VAE with manual annealing in either direction serves both baseline and inspiration for our work.

Our motivation for studying VAE is to generate fake yet realistic test data. Such data has a wide range of applications including testing systems involving input validation, performance testing, and UI design testing. We are particularly interested in generating samples that respect the correlations among multiple fields/columns of the training data, and we would like our generative model to discover and learn such correlations in an unsupervised fashion. Generating such fake yet realistic data is beyond the capability of a simple fuzzer, and to our knowledge such correlation is rarely measured independently from the reconstruction loss in the existing literature. In the following sections, we will first provide further background on generalized $\beta$-VAE. We will then describe the benchmark data set, followed by the tree recursive model generated by our framework and the baseline generation quality achievable with a generalized $\beta$-VAE. With the stage set, we keep the encoder-decoder pair constant and proceed to diagnose why the model fails to capture the full extent of the field correlations. First, we measure what the total VAE loss would be in our models for generated samples as if they were training or testing data, and we term it generated loss. We find that generated loss lags behind or even increases during the training process, in comparison to the training/testing loss. We believe that elevated generated loss indicates that information about the training examples is not diffused properly in the latent space, either due to overlapping posterior distributions or holes in the latent space. Motivated by this discovery, we seek improved variational methods more adaptive to local distribution of mean latent vector of training examples and capable of diffusing information throughout the latent space. Finally, we demonstrate that augmenting training data with generated variants under small $\beta$ (augmented training), and training a VAE with multiple values of $\beta$ simultaneously (multiscale VAE) are such variational methods and are closely related.

Our main contributions are as follows: 1) We propose generated loss, the total VAE loss of generated samples, as a diagnostic metric of generation quality of VAEs. 2) We propose augmented training, augmenting training data with generated variants, as a variational method for training VAEs to achieve superior generation quality. 3) Alternatively, we propose multiscale VAE, a VAE trained with multiple $\beta$ values simultaneously which is more tunable, captures aggregated characteristics like correlations more accurately, but tends to encode less details.

\section{Background}

\subsection{Generalized \texorpdfstring{$\beta$}{beta}-VAE}

Neural network-based autoencoders have long been used for unsupervised learning \citep{ballard1987modular} and variations like denoising autoencoder have been proposed to learn a more robust representation \citep{Vincent:2010:SDA:1756006.1953039}. The use of autoencoder as a generative model, however, only took off after the invention of VAE \citep{kingma2013auto}, which is trained to maximize the evidence lower bound (ELBO) of the log-likelihood of training examples $x$

\begin{equation}
\log p(x) \ge \mathbb{E}[\log p_\theta(x|z)] - KL(q_\lambda(z|x)||p(z))
\end{equation}

where $KL(\cdot||\cdot)$ is the KL-divergence between two distributions and $z$ is the latent vector, whose prior distribution $p(z)$ is most commonly assumed to be multivariate unit Gaussian. $\log p_\theta(x|z)$ is given by the decoder, and $q_\lambda(z|x)$ is the posterior distribution of the latent vector given by the stochastic encoder, whose operation can be made differentiable through the reparameterization trick $z = \mu_\lambda(x) + \sigma_\lambda(x) \odot \epsilon$, $\epsilon \sim \mathcal{N}(0, 1)$ if $q_\lambda(z|x)$ is assumed to be a diagonal-covariance Gaussian.

A common modification to the ELBO of VAE is to add a hyperparameter $\beta$ to the KL-divergence term and use the following objective function:

\begin{equation}
\mathbb{E}[\log p_\theta(x|z)] - \beta KL(q_\lambda(z|x)||p(z)) \label{eq:beta_vae}
\end{equation}

where $\beta$ controls the strength of the information bottleneck on the latent vector. For higher values of $\beta$, we accept lossier reconstruction, in exchange of higher effective compression ratio. This hyperparameter $\beta$ has been theoretically justified as a KKT multiplier for maximizing $\mathbb{E}[\log p_\theta(x|z)]$ under the inequality constraint that the KL-divergence must be less than a constant \citep{higgins2017beta}. In practice, $\beta$ is usually kept constant \citep{higgins2017beta} or manually annealed to increase over time \citep{DBLP:journals/corr/BowmanVVDJB15,sonderby2016ladder,DBLP:journals/corr/abs-1804-02135}.

In both cases, the generator $g(z)$ samples from the probability distribution given by the decoder $p_\theta(\tilde{x}|z)$ where $z$ is a random latent vector in generation time:

\begin{equation}
\tilde{x} = g(z) \sim p_\theta(\tilde{x}|z)
\end{equation}

\subsection{Benchmark data set and metric}
\label{sec:benchmark_data}

Addresses are a frequently encountered data type here at Google. It is a simple data type, but features intuitive yet non-trivial correlations among fields. Such correlation is perhaps easy to capture for specifically designed classifiers and regressors, but it is far harder to train generative models to generate samples that respect such correlation in an unsupervised fashion. Therefore, an address data set can serve as a context-relevant benchmark data set for our framework for training structured data VAEs. Specifically, the OpenAddresses Vermont state data set is chosen for its moderate size (See Appendix \ref{appendix:vermont_state_address_data_set} for more details).

We focus on the correlation between zip (postal) code and coordinates (latitude, longitude) as an example of field correlations. We estimate the distribution of coordinates of addresses in a given zip code from the training examples, and use p-value as the metric for generated samples. Recall that p-value is defined as the probability that the given sample is more likely than another sample from the same distribution, given null hypothesis. In our case, we would like the null hypothesis to be true -- i.e., training examples and generated samples follow the same distribution. For a perfect model, p-values of the generated samples follow uniform distribution between 0 and 1.

In practice, we make the simplifying assumption that the coordinates follow 2-dimensional Gaussian distribution for addresses of a given zip code. We consider the zip code categorical variable and calculate the mean $\mu$ and the sample correlation matrix $\Sigma$ of the coordinates in the zip code. We can then apply the multivariate version of the two-tailed t test to determine whether generated coordinates $x$ in the zip code follow the same distribution:

\begin{align*}
d_m^2 &= (x - \mu)^T \Sigma^{-1} (x - \mu) \\
p &= 1 - \operatorname{CDF}_{\chi_k^2}(d_m^2)
\end{align*}

where $d_m^2$ is the Mahalanobis distance squared, $\operatorname{CDF}_{\chi_k^2}$ is the cumulative distribution function (CDF) of chi-squared distribution with $k=2$ degrees of freedom.

\subsection{Tree recursive model}

Our address model consists of encoder-decoder modules. The latent space is 128-dim so all encoder-decoder modules produce and consume 128-dim vectors. The string fields (street number, street name, unit, city, district, region, and zip code) are modelled by a shared seq-to-seq char-rnn \texttt{StringLiteral} module, whereas the two float fields (latitude and longitude) are collected and modelled jointly by the \texttt{ScalarTuple} module. The full address data is then modelled by the \texttt{Tuple} module, whose encoder RNN consumes the embedding vectors generated by the encoders of these child modules and whose decoder RNN generates embedding vectors to be decoded by the decoders of these child modules. The reconstruction loss term for each field is given equal weight as follows:

\begin{enumerate}
  \item String field loss: calculated as cross-entropy loss per character in nat, given by the \texttt{StringLiteral} decoder. Each string field is given equal weight 1.0 regardless of the length of the string, so characters in a shorter string are given more weight than ones in a longer string. The \texttt{StringLiteral} decoder implements scheduled sampling \citep{DBLP:journals/corr/BengioVJS15} and can be trained with character input drawn from its own softmax distribution (always sampling, AS), ground-truth characters of the training example (teacher forcing, TF), or arbitrary scheduled sampling (SS) in between.
  \item Float field loss: The \texttt{ScalarTuple} module models latitude and longitude jointly and performs PCA-whitening as a preprocessing step on the fly with moving mean vector and covariance matrix. The decoder network then tries to predict the 2 resultant zero-mean unit-variance values, with mean squared errors as the loss function.
  \item Skew loss: The \texttt{Tuple} decoder adds a special loss term dubbed skew loss, the mean squared error between the embedding generated by itself and the embedding given by the respective child encoder. It is given equal weight as the child module's reconstruction loss, experimentally found to help stabilizing the training process, and makes sure that child encoder and decoder use the same representation. The \texttt{Tuple} decoder performs autoregression on its own output and implements scheduled sampling, where the embedding given by the child encoder is considered the ground-truth and using the embedding generated by the \texttt{Tuple} decoder itself is considered `sampling'.
\end{enumerate}

The latent loss is the standard KL-divergence loss between 128-dim unit Gaussian and diagonal-covariance Gaussian. Since we use weighted average for the reconstruction loss, we consider KL-divergence per latent dimension the latent loss and report its relative weight as $\beta$ in Eq \eqref{eq:beta_vae}.

The encoders of our framework only produces an embedding vector. In order to train a VAE, we interpret the embedding vector as the mean vector $\mu$ and generate the standard deviation vector $\sigma(\mu)$ from it with a standard deviation network. Our justification is that the embedding vector of a generative model should contain all the relevant information about the example, and this design simplifies the modular architecture. We did also test a more conventional architecture that generates both $\mu$ and $\sigma$ from the last layer on equal footing but found no qualitative difference in the model's behaviors. For more implementation details, see Appendix \ref{appendix:tree_recursive_model_details}.

\subsection{Training and generation}

Throughout experiments reported in this paper, the model is trained end-to-end using the Adam optimizer \citep{DBLP:journals/corr/KingmaB14} with initial learning rate $2.5 \times 10^{-4}$ and the rest of the TensorFlow default $\beta_1 = 0.9, \beta_2=0.999, \epsilon=10^{-8}$. The learning rate decays continuously by a factor of 0.99 per 1000 steps, and the gradients are clipped by the L2 global norm at 0.01. The experiments have a fixed budget of 2M training steps with batch size 256, running on 32 workers unless indicated otherwise. When KL-divergence warm-up and/or scheduled sampling are used, they have the same warm-up period with linear schedule.

With our focus on the difference between generated samples and training/testing examples, we do not want their difference to be trivially attributed to the difference in mean or covariance of their latent space distributions. Therefore, we sample from the multivariate Gaussian distribution closest to the distribution of the sampled latent vectors of the training data instead of the stronger assumption that they follow the unit Gaussian distribution. That is, we keep track of the moving mean $\mu$ and the moving covariance matrix $\Sigma$ of the sampled latent vectors during the training process, and sample from $\mathcal{N}(\mu, \Sigma)$ for generation. To assess the generation quality of trained models, we measure the p-values of generated coordinates in the generated zip code for 10000 generated samples. As described in described in Sec~\ref{sec:benchmark_data}, their ideal distribution is uniform distribution between 0 and 1, with mean = median = 0.5, standard deviation = $\frac{1}{\sqrt{12}}$. If the generated zip code is not found in the training data, the p-value is considered 0. Other than the p-values, we subjectively inspect the street names of generated samples, interpolation between training examples on the map, and measure the average Levenshtein distance per character $\bar{d}_{\mathrm{Levenshtein}}$ between the original street name and its reconstruction as a proxy for how much detail is encoded. Since we divide the Levenshtein distance by the length of the original street name, $\bar{d}_{\mathrm{Levenshtein}} \le 1$ as long as the reconstructed street name is not longer than the original. We measure the average Levenshtein distance per character $\bar{d}_{\mathrm{Levenshtein}}$ for each model over 10000 training examples that are randomly selected each time.

\section{\texorpdfstring{$\beta$}{beta}-VAE baseline}

Here we report the generation quality of baseline generalized $\beta$-VAE, measured by the p-values of generated samples and $\bar{d}_{\mathrm{Levenshtein}}$ of reconstructed training examples. These experiments use the full 2M steps as the warm-up period, and the ground-truth probability decreases linearly from 1 to 0 for scheduled sampling experiments. For a rough measure of the reproducibility, we rerun the best experiments with the same hyperparameters.

\begin{table}[ht]
  \centering
  \caption{$\beta$-VAE baseline performance}\label{tab:baseline_digest}
  \begin{tabular}{| c | c | c | c | c |}
    \hline
    & mean & median & stddev & $\bar{d}_{\mathrm{Levenshtein}}$ \\ %
    \hline \hline
    Tuple SS + String TF & 0.246 -- \textbf{0.261} & 0.0450 -- \textbf{0.0606} & 0.321 -- 0.329 & 0.114 -- 0.111 \\ %
    \hline
    Tuple AS + String TF & 0.240 -- 0.249 & 0.0448 -- 0.0505 & 0.317 -- 0.322 & 0.184 -- 0.149 \\ %
    \hline
    Always Sampling & 0.179 -- 0.203 & < 0.01 & 0.293 -- 0.303 & 0.0234 -- 0.0511 \\ %
    \hline
    Scheduled Sampling & 0.215 -- 0.247 & < 0.01 -- 0.0237 & 0.317 -- 0.331 & 0.0865 -- 0.0974 \\ %
    \hline
    Teacher Forcing & 0.0178 & 0 & 0.0970 & 0.0961 \\ %
    \hline
  \end{tabular}
\end{table}

Teacher forcing for strings makes generated street names more realistic, even though sampling for strings seems to drive down $\bar{d}_{\mathrm{Levenshtein}}$. What is happening is that sampling forces the string model to mindlessly generate the exact nth letter at the nth position, regardless of which letters are generated previously. This results in reconstructions such as ``PAINT WORKS RD'' $\to$ ``PAINT POIKS RD'' and nonsensical generated street names such as ``LINDINS PNON  ''. Sampling for Tuple is essential for the model to capture the correlations between fields, and scheduled sampling seems to hold a slight edge over always sampling by starting with an information shortcut directly from the child encoder to the Tuple RNN (See also Fig~\ref{fig:generated_loss_bpc}). We find that models trained with fixed $\beta$ often experience training failure characterized by increasing KL-divergence and elevated bits per character (BPC) during the training process relative to the KL-divergence warm-up counterpart. We hypothesize that the model has difficulty performing autoregression to take advantage of the autocorrelations in the presence of persistent noise due to latent vector sampling. Interpolation between two training examples by a model trained with the Tuple SS + String TF scheme tends to be a straight line on the map. Even though it bends for nearby population centers, the interpolation still passes through multiple sparsely populated areas like state forests, where there are few addresses. The model seems to recognize that city and zip code are categorical variables, but it indiscriminately tries to interpolate street name and number, even though these two are details of the training examples and may not make sense to interpolate. In the shown example, the model makes up multiple addresses that start with ``147, HARTS RD, GROTON'' due to a training example nearby that starts with ``147, HARTS RD, TOPSHAM''.

Simple char-rnn VAEs trained on concatenated string expression of the training examples can actually generate good samples in terms of zip-coordinate correlations, but occasionally generate malformed samples that result in errors when converted back to structured data (Appendix \ref{appendix:text_model}). Simpler multi-seq-to-multi-seq model without autoregression at the \texttt{Tuple} level fails to capture any of the correlations (Appendix \ref{appendix:pass_through_model}) and neither do generative model frameworks that regularize global latent vector distributions but not continuity of the decoder like adversarial autoencoder (Appendix \ref{appendix:adversarial_autoencoder_model}) and Wasserstein autoencoder (Appendix \ref{appendix:wasserstein_autoencoder_model}).

\section{Generated loss}

\begin{figure}[ht]
\centering
\caption{Loss (left) and BPC (right) for training data, testing data, and generated samples during KL-divergence warm-up ($\beta_{\mathrm{start}} = 0$, $\beta_{\mathrm{end}} = 0.384$), Tuple SS + String TF. KL-divergence warm-up drives increase in training and testing loss. Due to KL-divergence warm-up, steady generated loss actually implies decreasing reconstruction loss for generated samples, as evidenced by the decreasing generated BPC.}\label{fig:warm_up_generated_loss}
\includegraphics[width=\textwidth]{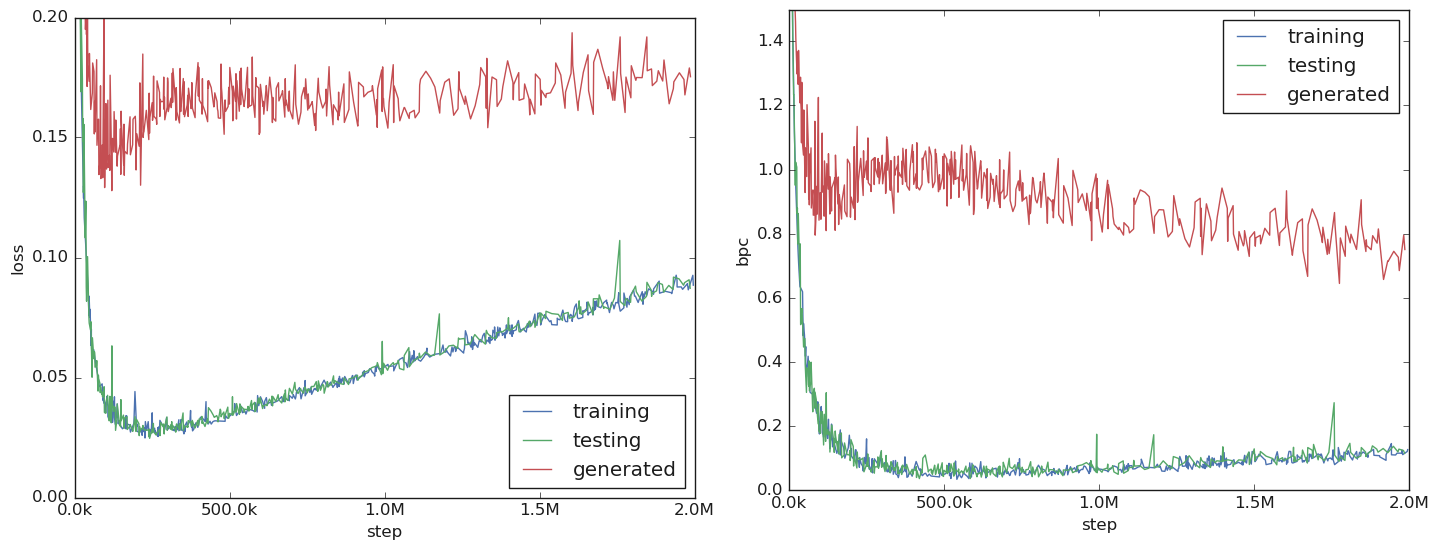}
\end{figure}

\begin{figure}[ht]
\centering
\caption{Loss (left) and BPC (right) for generated samples during the training process with the same KL-divergence warm-up under different training schemes, generated loss of the Tuple SS + String TF scheme is the same as that of Fig~\ref{fig:warm_up_generated_loss}.}\label{fig:generated_loss_bpc}
\includegraphics[width=\textwidth]{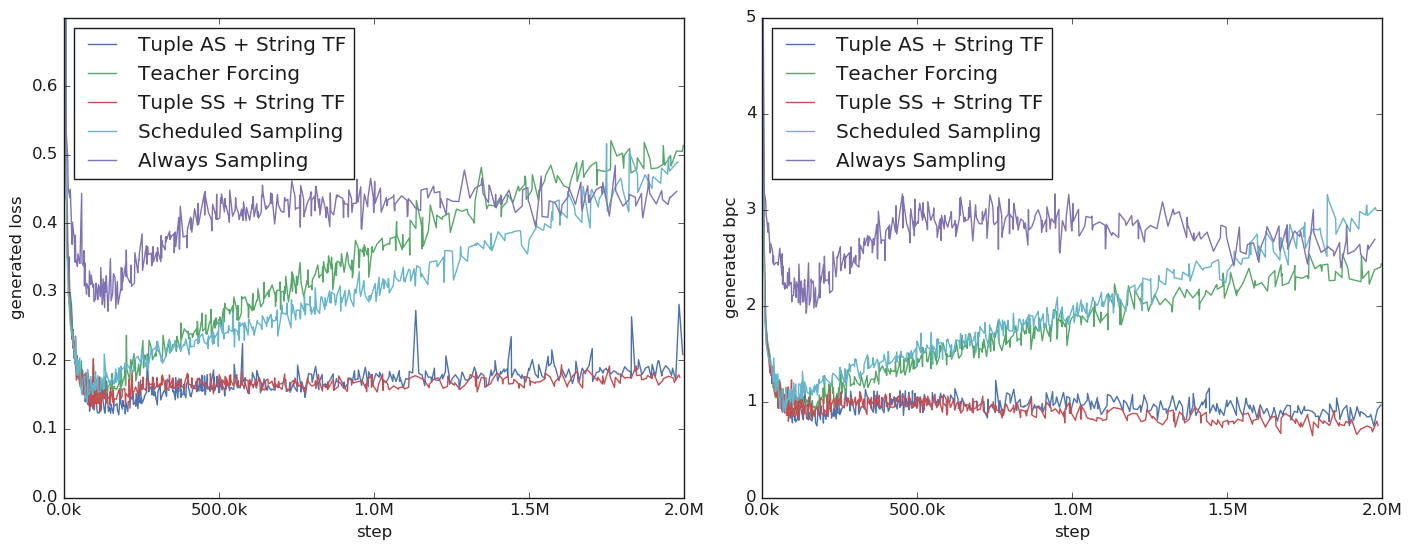}
\end{figure}

For these Vermont state address models, we never observe overfitting. That is, training loss and testing loss always change in sync and the model is indeed maximizing the log-likelihood of the data from the true distribution. This is not the case, however, for samples generated by the model. We measured what the VAE loss would be in our models for generated samples as if they were training or testing data, and we term it generated loss. We found that generated loss lags behind under the Tuple sampling + String TF schemes, and actually increases under all of the other training schemes during the training process (Fig~\ref{fig:warm_up_generated_loss} and~\ref{fig:generated_loss_bpc}).

Except for the Tuple sampling + String TF training schemes, generated loss actually increases during the training process, so the model is not maximizing the log-likelihood of its generated samples by its own estimate. In other words, $\beta$-VAE fails to establish bijection between the latent space and the data space for generated samples not from the true distribution. Perhaps we should not find this surprising: in the training process of VAE, we minimize the reconstruction loss from a latent vector sampled from a distribution centered around the mean latent vector. So if we start from a generated sample that is mapped into the neighborhood of a training example in the latent space by the encoder, encoding using the mean vector followed by decoding will result in a sample more similar to the said training example. Indeed, we observe that p-values of generated samples tend to increase over repeated encoding and decoding. What is more surprising is that even the training examples themselves are not immune to this. In fact, they seem to converge faster but to the same distribution of p-values as the generated samples, after just one round of encoding / decoding. Apparently, the gravitational pull of training examples is additive and exerting on themselves. Since the latent vectors during training are sampled from a Gaussian distribution, this gravitational pull should diminish exponentially as the distance in the latent space increases, perhaps not unlike gravity modified with a Yukawa-type potential. For models that encode the street names, the street names of generated samples also tend to converge to real street names in the training data after repeated encoding and decoding. For detailed formalism and plots, see Appendix \ref{appendix:repeated_encoding_decoding}.

\section{Augmented training}

Our results of generated loss measurements suggest that information from the training data isn't sufficiently propagated. So we propose the following scheme to facilitate biased diffusion in the latent space by training on generated variants.

\begin{adjustbox}{center}
\framebox{
\parbox{\textwidth}{
After \texttt{gen\_start\_step} steps:
\begin{enumerate}
  \item Initialize $n_{\mathrm{augmented}}$ augmented latent vectors with sampled latent vectors of the current training batch.
  \item Augment next training batch with variants generated from the augmented latent vectors.
  \item After a training step, each augmented latent vector is replaced with either:
  \begin{enumerate}
    \item The sampled latent vector of an example from the current training batch, selected without replacement, with probability $p_{\mathrm{sampled}}$.
    \item The sampled latent vector of the variant generated from it with probability $(1-p_{\mathrm{sampled}})$.
  \end{enumerate}
  \item Repeat from 2.
\end{enumerate}
}
}
\end{adjustbox}

Intuitively, augmented training extends the standard VAE training scheme. Instead of just taking one `hop' from the mean vector of a training example and minimizing the reconstruction loss from the sampled latent vector, we actually generate a reconstruction from the sampled latent vector, run it through the encoder, take a second `hop' from the mean vector of the reconstruction and minimize the reconstruction loss from the augmented latent vector to the reconstruction, and so on. The augmented latent vector is initialized from the sampled latent vector of a training example, and before we run generation from it the model was just trained to minimize reconstruction loss from it. Therefore, the reconstruction generated from the sampled latent vector is likely to be similar to the original training example. The similarity will decay over repeated encoding / decoding due to the model's capacity limit and the noise introduced by latent vector sampling, so we re-initialize it with probability $p_{\mathrm{sampled}}$ such that the average lifetime is $\frac{1}{p_{\mathrm{sampled}}}$ steps, which turned out to be 5 steps for the optimized experiments below. We only start augmented training after \texttt{gen\_start\_step} steps to make sure that the model is ready to generate reasonable reconstructions, and $n_{\mathrm{augmented}}$ controls the number of augmented latent vectors we use. Formally, we train the model with reconstructions from the following sequence in addition to the training examples $x$:

\begin{align*}
x' = g(z), z &\sim \mathcal{N}(\mu_\lambda(x), \sigma_\lambda(x)) \\
x'' = g(z'), z' &\sim \mathcal{N}(\mu_\lambda(x'), \sigma_\lambda(x'))  \\
&\dots \\
x^{(n)} = g(z^{(n-1)}), z^{(n-1)} &\sim \mathcal{N}(\mu_\lambda(x^{(n-1)}), \sigma_\lambda(x^{(n-1)}))  \text{ for } n > 0 \\
\end{align*}

In terms of objective function, we have

\begin{equation}
\label{eq:augmented}
\sum_{n=0}^{\infty} p^{\min(n, 1)}_{\mathrm{sampled}} (1 - p_{\mathrm{sampled}})^{\max(n-1, 0)} (\mathbb{E}[\log p_\theta(x^{(n)}|z^{(n)})] - \beta KL(q_\lambda(z^{(n)}|x^{(n)})||p(z^{(n)})))
\end{equation}

Assuming that $n_{\mathrm{augmented}}$ is equal to the training batch size, as is the case for our experiments. It is worth pointing out that $z^{(n)}$ are sampled latent vectors instead of the mean vector given by the encoder in the previous section. Our experiments showed that augmented training does not improve generation quality without such noise injection.

\begin{table}[ht]
\centering
  \caption{Augmented training performance}\label{tab:augmented_digest}
  \begin{tabular}{| c | c | c | c | c |}
    \hline
    & mean & median & stddev & $\bar{d}_{\mathrm{Levenshtein}}$ \\ %
    \hline
    Tuple SS + String TF & 0.401 -- \textbf{0.401} & 0.321 -- \textbf{0.324} & 0.373 -- 0.376 & 0.239 -- 0.195 \\ %
    \hline
    Scheduled Sampling & 0.317 -- 0.335 & 0.137 -- 0.180 & 0.354 -- 0.358 & 0.116 -- 0.115 \\ %
    \hline
  \end{tabular}
\end{table}

\begin{figure}[ht]
\centering
\caption{Loss (left) and BPC (right) for training data, testing data, and generated samples during augmented training. Generated loss/BPC now move in sync with their training/testing counterparts. Unusually, training loss/BPC are higher than their testing counterparts since training batches are augmented with generated variants, which are not from the true distribution exactly and therefore more challenging for the model. Loss is mostly driven by KL-divergence warm-up followed by cool-down. Due to build optimization issues, we had to interrupt the training process twice between 500k and 1M steps.}\label{fig:augmented_generated_loss}
\includegraphics[width=\textwidth]{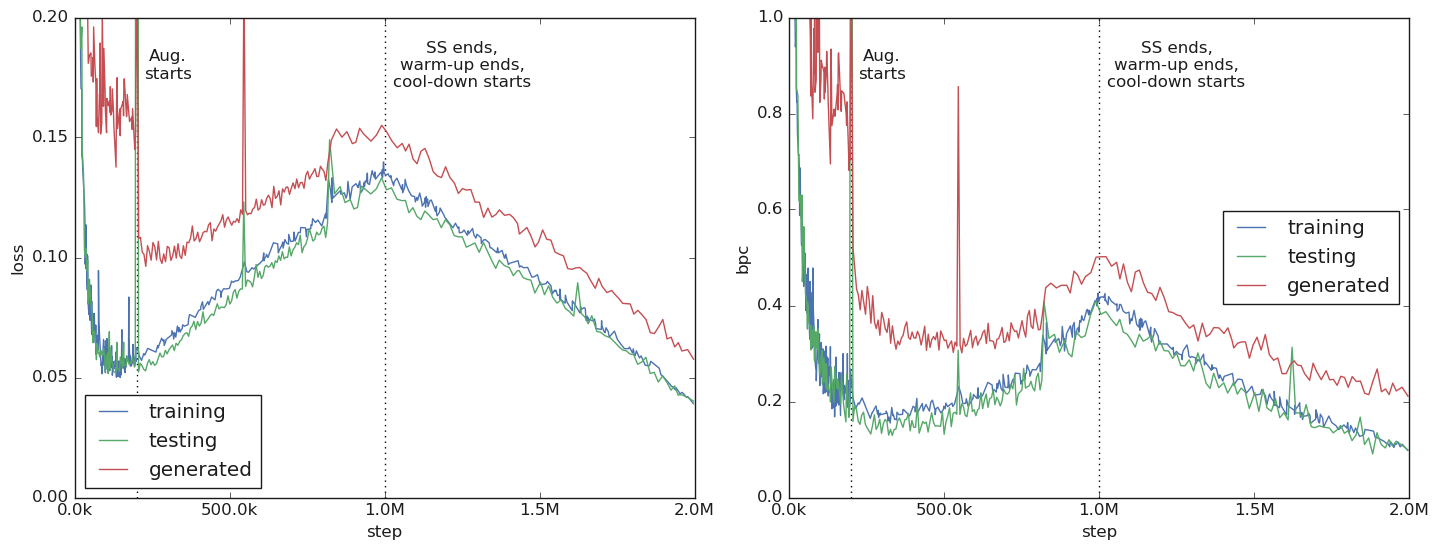} %
\end{figure}

We can see that augmented training improves model's generation quality and generated loss (Fig~\ref{fig:augmented_generated_loss}, taken from the run marked by the bold font in Table~\ref{tab:augmented}). Reduced generated loss indicates better embedding of the generated samples, even though it is still not as low as training/testing loss. KL-divergence warm-up followed by cool-down outperforms simple warm-up, despite identical $\beta_{\mathrm{end}}$. We suspect that with simple KL-divergence warm-up the difference between real and fake data gets more entrenched so it is harder for augmented latent vector to escape their potential well, following the gravity analogy.

As part of our observation for the model's interpolation, we find that an augmented training model settles more often on a generated street name instead of fully interpolating the street names of training examples. For example, in a interpolation between training examples with street names ``HARTS RD'' and ``SECOND ST'', the most common street name given is actually ``S   MAIN ST''. The theme of non-linearity continues as we plot the interpolation on the map, which twists and turns to avoid sparsely populated areas like state forests and goes through population centers like cities and towns instead.

\section{Multiscale VAE}
\label{sec:multiscale}

The fact that we can improve the model with KL-divergence warm-up and cool-down indicates that it does not take many steps to train the standard deviation network. We also make the observation that objective functions with different $\beta$ are not necessarily in conflict with each other. Training with $\beta_{\mathrm{max}}$, the highest possible $\beta$ before $q_\lambda(z|x)$ collapses into multivariate unit Gaussian optimizes the global structure of the latent space. Training with smaller $\beta$ optimizes the local structure, and training with $\beta=0$ optimizes autoencoding. Therefore, we propose the following training scheme.

\begin{adjustbox}{center}
\framebox{
\parbox{\textwidth}{
\begin{enumerate}
  \item Initialize $n_{\mathrm{KL\_weight}}$ standard deviation networks $\sigma_{\lambda, i}(\mu)$, where each standard deviation network is associated with a distinct but constant $\beta$ value $\beta_i$. Without loss of generality, we assume $\beta_i < \beta_{i+1}$.
  \item Assign $(\beta_i, \sigma_{\lambda, i}(\mu))$ pairs evenly to workers, which otherwise share the same encoder / decoder.
  \item Train the model with these workers.
\end{enumerate}
}
}
\end{adjustbox}

In terms of objective function, we have

\begin{equation}
\label{eq:multiscale}
\begin{aligned}
\sum_{i=0}^{\mathrm{kl\_weight\_levels} - 1} \mathbb{E}[\log p_\theta(x|z^{(i)})] &- \beta_i KL(q_{\lambda, i}(z^{(i)}|x)||p(z^{(i)})) \\
\text{where } z^{(i)} &\sim \mathcal{N}(\mu_\lambda(x), \sigma_{\lambda, i}(\mu_\lambda(x)))
\end{aligned}
\end{equation}

It is tempting to draw connections and contrast between this multiscale objective function and the augmented objective function Eq \eqref{eq:augmented}. Intuitively, augmented latent vectors $z^{(n)}$ should get further and further away from the mean vector $\mu_\lambda(x)$ in average after more and more `hops', and indeed at the limit of perfect encoder / decoder $\mu_\lambda(g(z)) \to z$, small $\beta \to 0$ and locally constant $\sigma_\lambda(x)$ in the neighborhood of $x$, $z^{(n)} \sim \mathcal{N}(\mu_\lambda(x), \sqrt{n+1} \sigma_\lambda(x))$ as sum of Gaussian random variables. Perhaps augmented training has similar effects on the model as multiscale VAE with geometrically spaced $\beta$ values where terms with higher $n$ in Eq \eqref{eq:augmented} serve the role of workers with higher $\beta_i$. For experiments partially motivated by this observation, see Appendix \ref{appendix:alternative_multiscale}. In this section, we set $n_{\mathrm{KL\_weight}} = 32$ and $\beta_0 = \frac{1}{32}\beta_{\mathrm{max}}, \beta_1 = \frac{2}{32}\beta_{\mathrm{max}}, \dots, \beta_{31} = \beta_{\mathrm{max}}$, which seem to have better behavior.

For the experiments below, no augmented training is used and they always employ scheduled sampling and KL-divergence cool-down for the first 1M training steps. For experiments combining this setup and augmented training, see Appendix \ref{appendix:multiscale_augmented}.

\begin{table}[ht]
\centering
  \caption{Multiscale VAE only performance}\label{tab:multiscale_digest}
  \begin{tabular}{| c | c | c | c | c |}
    \hline
    & mean & median & stddev & $\bar{d}_{\mathrm{Levenshtein}}$ \\ %
    \hline
    Tuple SS + String TF & 0.476 -- \textbf{0.509} & \textbf{0.494} -- 0.580 & 0.382 -- 0.384 & 0.916 -- 0.921 \\ %
    \hline
    Scheduled Sampling & 0.402 -- 0.411 & 0.349 -- 0.361 & 0.367 -- 0.371 & 0.497 -- 0.525 \\ %
    \hline
  \end{tabular}
\end{table}

We can see that multiscale VAE alone outperforms even augmented training models in terms of generation quality. In fact, it is no longer obvious which model is the best. With optimized hyperparameters, one run results in a model that generates samples whose p-values are slightly below that of training examples, and the other results in a model that generates samples whose p-values are slightly above (the latter one is arbitrarily chosen for the following figures). The optimal value of $\beta_{\mathrm{max}}$ in our case is in the range 0.64 -- 1.28, consistent with our results with $\beta$-VAE. Since multiscale VAE by design already optimizes the log-likelihood of training examples for $\beta$ values in the interval $(0, \beta_{\mathrm{max}}]$, KL-divergence warm-up from 0 unsurprisingly doesn't help. However, modest KL-divergence cool-down seems to be beneficial and improves the robustness of model performance w.r.t. hyperparameter tuning of $\beta_{\mathrm{max}}$.

\begin{figure}[ht]
\centering
\caption{Loss (left) and BPC (right) for training data, testing data, and generated samples during the training process of multiscale VAE only on the worker with the lowest $\beta$ value $\beta_0 = \frac{1}{32}\beta_{\mathrm{max}}$.}\label{fig:multiscale_generated_loss}
\includegraphics[width=\textwidth]{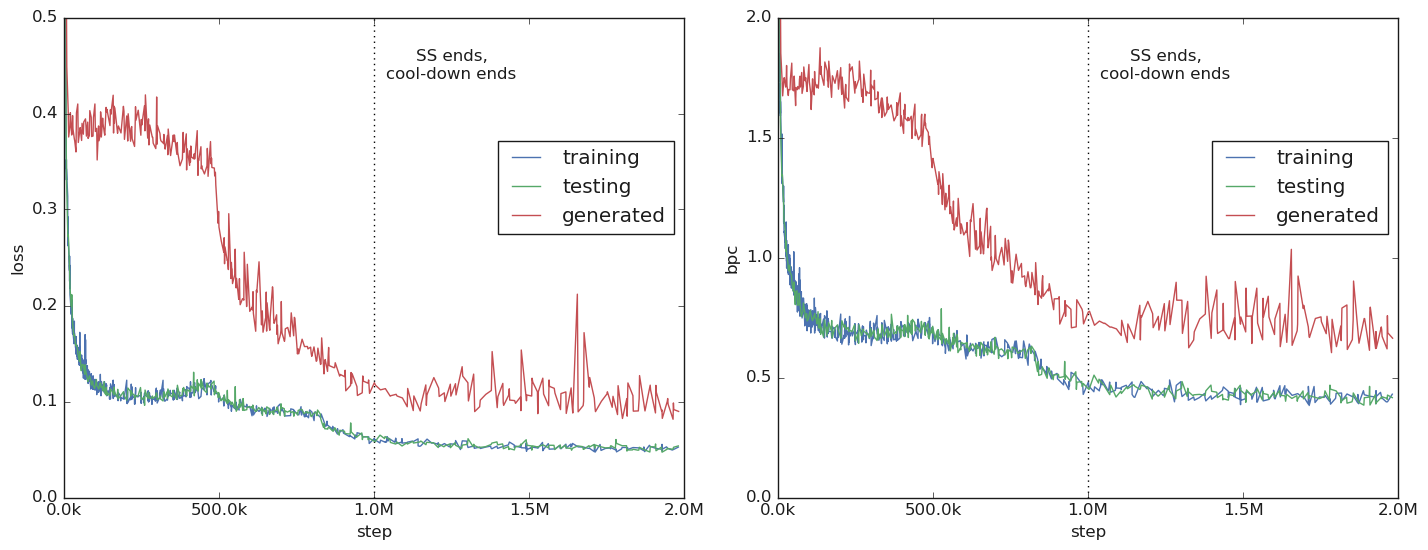}
\end{figure}

While not designed to do so, optimized multiscale VAE alone does have lower generated loss (Fig~\ref{fig:multiscale_generated_loss}) than the $\beta$-VAE baseline. While the multiscale VAE model still reconstructs and interpolates between street numbers, it no longer does so for street names. Instead, it makes up street names by autoregression, in a sense exhibiting partial posterior collapse. This behavior is intuitively sensible: The model sees a wide variety of street names in close proximity of each other and associated with the same city and zip code, and subsequently concludes that street names are details not to memorize for each individual training example. In aggregation, however, the multiscale VAE model actually generates more samples with street names from the training data than the optimized $\beta$-VAE model (60\% vs. 44\%). When the interpolation given by a Multiscale VAE is plotted on the map, it snakes through multiple population centers and tries to stay within their neighborhoods as long as possible to minimize coordinate and zip code loss terms. In the optimal case, the interpolation adapts the property of a space-filling curve. Both tendencies are present in the augmented training models, but even stronger for multiscale VAE.

\section{Conclusion}

We described the Vermont state address benchmark data set, a field correlation metric used to quantify generation quality, and our discrete VAE based on a generated tree recursive model. We showed that even when trained with KL-divergence warm-up and scheduled sampling, generalized $\beta$-VAE only demonstrates limited capacity in capturing such field correlations, and most of the issue is with the variational method instead of the encoder-decoder pair. More specifically,

\begin{enumerate}
  \item VAE loss of generated samples (generated loss) may lag behind or even increase during the training process and serves as a useful metric for VAE optimization. The model tends to make mistakes in the direction of typical training examples, even for the training examples themselves.
  \item Both generation quality and generated loss can be improved by augmenting training data with generated variants (augmented training).
  \item Training VAE with multiple $\beta$ values and standard deviation networks simultaneously (multiscale VAE) is a formally related, tunable technique. The resulted model tends to encode less details but offer superior generation quality in terms of aggregated properties like field correlations.
\end{enumerate}

Admittedly, we have not fully solved the issues observed in our work, and it is speculative whether we have hit upon certain fundamental limitations of the VAE framework. For early results of applying these ideas to an image VAE, see \citet{anon2019mnist}.

\subsubsection*{Author contributions}

J.C. contributed the idea and implementation of generated loss measurement, augmented training, multiscale VAE, and the use of p-value as correlation metric. J.C. also implemented the Alala framework in collaboration with DeLesley Hutchins, with a focus on the decoders, VAE training scheme, and the engine for model generation from Protocol Buffer message definition. G.H. contributed the idea of using the Vermont state address data set and correlations of generated data as the generation quality metric. G.H. also implemented the trainer binary for the Vermont state address model and the Python module for measuring correlations.

\subsubsection*{Acknowledgments}

DeLesley Hutchins designed and implemented the modular encoder-decoder architecture and the bidirectional RNN \texttt{Tuple} encoder for a different purpose, and both are incorporated into the Alala framework. DeLesley Hutchins also first suggested the use of scheduled sampling and pass-through baseline model and provided valuable critiques to an early draft of this paper. We would also like to thank Irina Higgins for extensive internal review and helpful feedbacks.

\bibliography{ms}

\begin{thebibliography}{22}
\providecommand{\natexlab}[1]{#1}
\providecommand{\url}[1]{\texttt{#1}}
\expandafter\ifx\csname urlstyle\endcsname\relax
  \providecommand{\doi}[1]{doi: #1}\else
  \providecommand{\doi}{doi: \begingroup \urlstyle{rm}\Url}\fi

\bibitem[Akuzawa et~al.(2018)Akuzawa, Iwasawa, and
  Matsuo]{DBLP:journals/corr/abs-1804-02135}
Kei Akuzawa, Yusuke Iwasawa, and Yutaka Matsuo.
\newblock Expressive speech synthesis via modeling expressions with variational
  autoencoder.
\newblock \emph{CoRR}, abs/1804.02135, 2018.
\newblock URL \url{http://arxiv.org/abs/1804.02135}.

\bibitem[Ballard(1987)]{ballard1987modular}
Dana~H Ballard.
\newblock Modular learning in neural networks.
\newblock In \emph{AAAI}, pp.\  279--284, 1987.

\bibitem[Barron(2017)]{DBLP:journals/corr/Barron17a}
Jonathan~T. Barron.
\newblock Continuously differentiable exponential linear units.
\newblock \emph{CoRR}, abs/1704.07483, 2017.
\newblock URL \url{http://arxiv.org/abs/1704.07483}.

\bibitem[Bengio et~al.(2015)Bengio, Vinyals, Jaitly, and
  Shazeer]{DBLP:journals/corr/BengioVJS15}
Samy Bengio, Oriol Vinyals, Navdeep Jaitly, and Noam Shazeer.
\newblock Scheduled sampling for sequence prediction with recurrent neural
  networks.
\newblock \emph{CoRR}, abs/1506.03099, 2015.
\newblock URL \url{http://arxiv.org/abs/1506.03099}.

\bibitem[Bowman et~al.(2015)Bowman, Vilnis, Vinyals, Dai, J{\'{o}}zefowicz, and
  Bengio]{DBLP:journals/corr/BowmanVVDJB15}
Samuel~R. Bowman, Luke Vilnis, Oriol Vinyals, Andrew~M. Dai, Rafal
  J{\'{o}}zefowicz, and Samy Bengio.
\newblock Generating sentences from a continuous space.
\newblock \emph{CoRR}, abs/1511.06349, 2015.
\newblock URL \url{http://arxiv.org/abs/1511.06349}.

\bibitem[Burgess et~al.(2018)Burgess, Higgins, Pal, Matthey, Watters,
  Desjardins, and Lerchner]{burgess2018understanding}
Christopher~P Burgess, Irina Higgins, Arka Pal, Loic Matthey, Nick Watters,
  Guillaume Desjardins, and Alexander Lerchner.
\newblock Understanding disentangling in $\beta$-vae.
\newblock \emph{arXiv preprint arXiv:1804.03599}, 2018.

\bibitem[Carlini et~al.(2018)Carlini, Liu, Kos, Erlingsson, and
  Song]{DBLP:journals/corr/abs-1802-08232}
Nicholas Carlini, Chang Liu, Jernej Kos, {\'{U}}lfar Erlingsson, and Dawn Song.
\newblock The secret sharer: Measuring unintended neural network memorization
  {\&} extracting secrets.
\newblock \emph{CoRR}, abs/1802.08232, 2018.
\newblock URL \url{http://arxiv.org/abs/1802.08232}.

\bibitem[Cho et~al.(2014)Cho, van Merrienboer, G{\"{u}}l{\c{c}}ehre, Bougares,
  Schwenk, and Bengio]{DBLP:journals/corr/ChoMGBSB14}
Kyunghyun Cho, Bart van Merrienboer, {\c{C}}aglar G{\"{u}}l{\c{c}}ehre, Fethi
  Bougares, Holger Schwenk, and Yoshua Bengio.
\newblock Learning phrase representations using {RNN} encoder-decoder for
  statistical machine translation.
\newblock \emph{CoRR}, abs/1406.1078, 2014.
\newblock URL \url{http://arxiv.org/abs/1406.1078}.

\bibitem[Chou(2019)]{anon2019mnist}
Jason Chou.
\newblock Generated loss and augmented training of mnist vae.
\newblock In \emph{arXiv}, 2019.

\bibitem[Goodfellow et~al.(2014)Goodfellow, Pouget-Abadie, Mirza, Xu,
  Warde-Farley, Ozair, Courville, and Bengio]{goodfellow2014generative}
Ian Goodfellow, Jean Pouget-Abadie, Mehdi Mirza, Bing Xu, David Warde-Farley,
  Sherjil Ozair, Aaron Courville, and Yoshua Bengio.
\newblock Generative adversarial nets.
\newblock In \emph{Advances in neural information processing systems}, pp.\
  2672--2680, 2014.

\bibitem[Higgins et~al.(2017)Higgins, Matthey, Pal, Burgess, Glorot, Botvinick,
  Mohamed, and Lerchner]{higgins2017beta}
Irina Higgins, Loic Matthey, Arka Pal, Christopher Burgess, Xavier Glorot,
  Matthew Botvinick, Shakir Mohamed, and Alexander Lerchner.
\newblock beta-vae: Learning basic visual concepts with a constrained
  variational framework.
\newblock In \emph{International Conference on Learning Representations}, 2017.

\bibitem[Ioffe \& Szegedy(2015)Ioffe and Szegedy]{DBLP:journals/corr/IoffeS15}
Sergey Ioffe and Christian Szegedy.
\newblock Batch normalization: Accelerating deep network training by reducing
  internal covariate shift.
\newblock \emph{CoRR}, abs/1502.03167, 2015.
\newblock URL \url{http://arxiv.org/abs/1502.03167}.

\bibitem[Kingma \& Ba(2014)Kingma and Ba]{DBLP:journals/corr/KingmaB14}
Diederik~P. Kingma and Jimmy Ba.
\newblock Adam: {A} method for stochastic optimization.
\newblock \emph{CoRR}, abs/1412.6980, 2014.
\newblock URL \url{http://arxiv.org/abs/1412.6980}.

\bibitem[Kingma \& Welling(2013)Kingma and Welling]{kingma2013auto}
Diederik~P Kingma and Max Welling.
\newblock Auto-encoding variational bayes.
\newblock \emph{arXiv preprint arXiv:1312.6114}, 2013.

\bibitem[Krizhevsky \& Hinton(2010)Krizhevsky and
  Hinton]{krizhevsky2010convolutional}
Alex Krizhevsky and Geoff Hinton.
\newblock Convolutional deep belief networks on cifar-10.
\newblock \emph{Unpublished manuscript}, 40\penalty0 (7), 2010.

\bibitem[Looks et~al.(2017)Looks, Herreshoff, Hutchins, and
  Norvig]{DBLP:journals/corr/LooksHHN17}
Moshe Looks, Marcello Herreshoff, DeLesley Hutchins, and Peter Norvig.
\newblock Deep learning with dynamic computation graphs.
\newblock \emph{CoRR}, abs/1702.02181, 2017.
\newblock URL \url{http://arxiv.org/abs/1702.02181}.

\bibitem[Makhzani et~al.(2015)Makhzani, Shlens, Jaitly, and
  Goodfellow]{DBLP:journals/corr/MakhzaniSJG15}
Alireza Makhzani, Jonathon Shlens, Navdeep Jaitly, and Ian~J. Goodfellow.
\newblock Adversarial autoencoders.
\newblock \emph{CoRR}, abs/1511.05644, 2015.
\newblock URL \url{http://arxiv.org/abs/1511.05644}.

\bibitem[Rezende \& Viola(2018)Rezende and Viola]{rezende2018taming}
Danilo~Jimenez Rezende and Fabio Viola.
\newblock Taming vaes.
\newblock \emph{arXiv preprint arXiv:1810.00597}, 2018.

\bibitem[Sajjadi et~al.(2018)Sajjadi, Bachem, Lucic, Bousquet, and
  Gelly]{sajjadi2018assessing}
Mehdi~SM Sajjadi, Olivier Bachem, Mario Lucic, Olivier Bousquet, and Sylvain
  Gelly.
\newblock Assessing generative models via precision and recall.
\newblock \emph{arXiv preprint arXiv:1806.00035}, 2018.

\bibitem[S{\o}nderby et~al.(2016)S{\o}nderby, Raiko, Maal{\o}e, S{\o}nderby,
  and Winther]{sonderby2016ladder}
Casper~Kaae S{\o}nderby, Tapani Raiko, Lars Maal{\o}e, S{\o}ren~Kaae
  S{\o}nderby, and Ole Winther.
\newblock Ladder variational autoencoders.
\newblock In \emph{Advances in neural information processing systems}, pp.\
  3738--3746, 2016.

\bibitem[{Tolstikhin} et~al.(2017){Tolstikhin}, {Bousquet}, {Gelly}, and
  {Schoelkopf}]{2017arXiv171101558T}
Ilya {Tolstikhin}, Olivier {Bousquet}, Sylvain {Gelly}, and Bernhard
  {Schoelkopf}.
\newblock {Wasserstein Auto-Encoders}.
\newblock \emph{arXiv e-prints}, art. arXiv:1711.01558, November 2017.

\bibitem[Vincent et~al.(2010)Vincent, Larochelle, Lajoie, Bengio, and
  Manzagol]{Vincent:2010:SDA:1756006.1953039}
Pascal Vincent, Hugo Larochelle, Isabelle Lajoie, Yoshua Bengio, and
  Pierre-Antoine Manzagol.
\newblock Stacked denoising autoencoders: Learning useful representations in a
  deep network with a local denoising criterion.
\newblock \emph{J. Mach. Learn. Res.}, 11:\penalty0 3371--3408, December 2010.
\newblock ISSN 1532-4435.
\newblock URL \url{http://dl.acm.org/citation.cfm?id=1756006.1953039}.

\end{thebibliography}
\bibliographystyle{ms}

\appendix
\section{Vermont state address data set}
\label{appendix:vermont_state_address_data_set}

A corpus of Vermont state addresses from the zip file \texttt{us\_northeast.zip} is downloaded from OpenAddresses. We decompressed it to use its \texttt{us/vt/statewide.csv} as the raw data. We then defined a simple Protocol Buffer message to represent its rows and for the purpose of model generation described in Appendix \ref{appendix:tree_recursive_model_details}:

\begin{verbatim}
message Address {
  optional float lat = 1;
  optional float long = 2;

  // We don't discount the possibility that some addresses could have
  // non-numerical entries for fields that seem like they should be numerical,
  // like street numbers for example:
  optional string number = 3;
  optional string street = 4;
  optional string unit = 5;
  optional string city = 6;
  optional string district = 7;
  optional string region = 8;
  optional string postcode = 9;
}
\end{verbatim}

We then split the data set into training, testing, and validation sets with 8:1:1 expected ratio. The training set contains 266450 examples, the testing set contains 33304 examples, and the validation set remains unused. In terms of total number of characters, the training set contains 7504720 characters among all string fields, or an average of 28.17 per training example. These sets used in our work can be downloaded from \url{https://github.com/EIFY/vermont\_address}. We have also trained models on different slices and found the results to be robust to slice change.

Regarding zip-coordinate correlations, we expect the p-values of coordinates given zip code to be uniformly distributed between 0 and 1 for the training set itself. As a sanity check, here are the stats of the p-values of the training set:

\begin{verbatim}
Mean: 0.521861141342
Median: 0.537469273433
Standard deviation: 0.298400
\end{verbatim}

Given finite training examples, these stats seem reasonable relative to the $n \to \infty$ limit mean = median = 0.5, standard deviation = $\frac{1}{\sqrt{12}}$.

\clearpage

\section{Tree recursive model implementation details}
\label{appendix:tree_recursive_model_details}

\begin{figure}
    \centering
    \caption{StringLiteral Module}
    \subfloat{{\includegraphics[width=\linewidth]{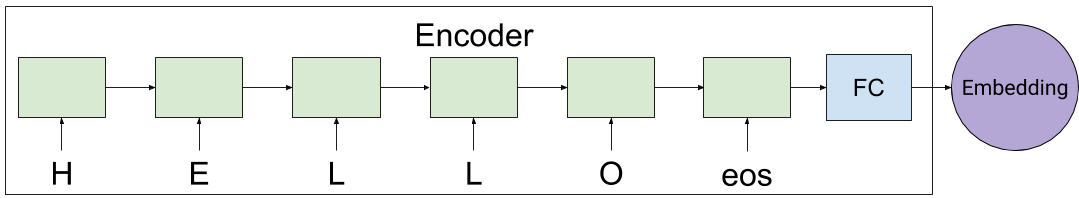}}}
    \qquad
    \subfloat{{\includegraphics[width=\linewidth]{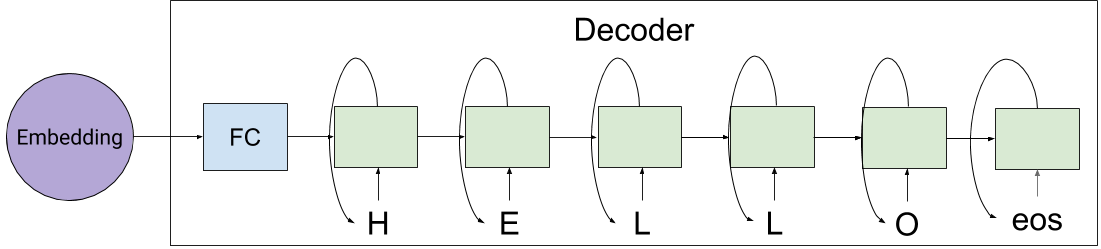}}}
    \label{fig:StringLiteral}
\end{figure}

\begin{figure}
    \centering
    \caption{ScalarTuple Module}
    \subfloat{{\includegraphics[width=\linewidth]{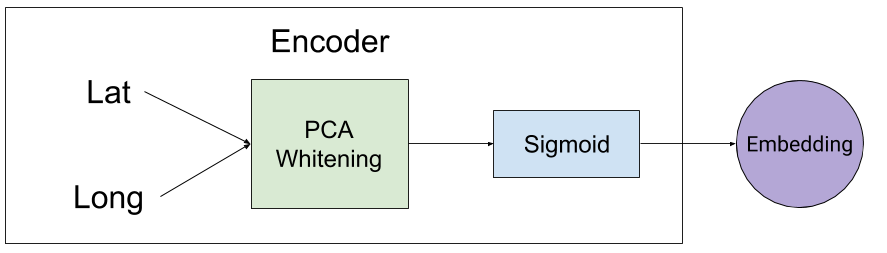}}}
    \qquad
    \subfloat{{\includegraphics[width=\linewidth]{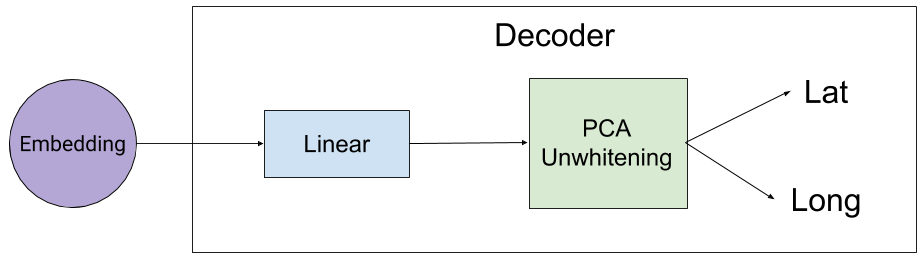}}}
    \label{fig:ScalarTuple}
\end{figure}

\begin{figure}
    \centering
    \caption{Tuple Module}
    \subfloat{{\includegraphics[width=\linewidth]{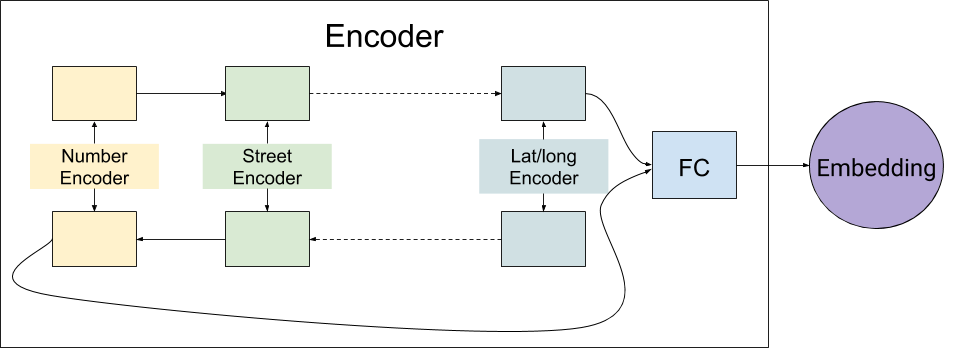}}}
    \qquad
    \subfloat{{\includegraphics[width=\linewidth]{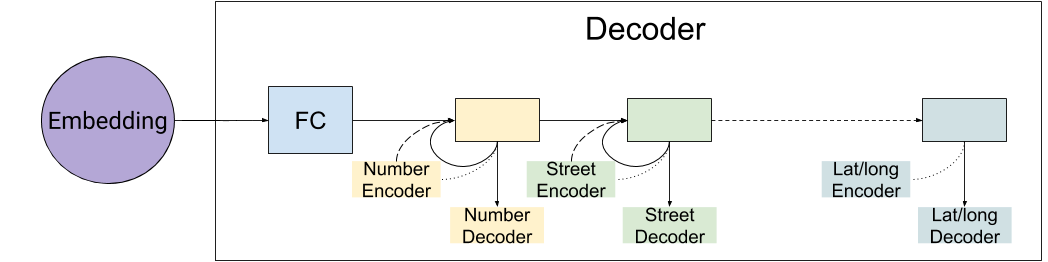}}}
    \label{fig:Tuple}
\end{figure}

\begin{figure}[ht]
\centering
\caption{Standard Deviation Network}
\includegraphics[scale=0.4]{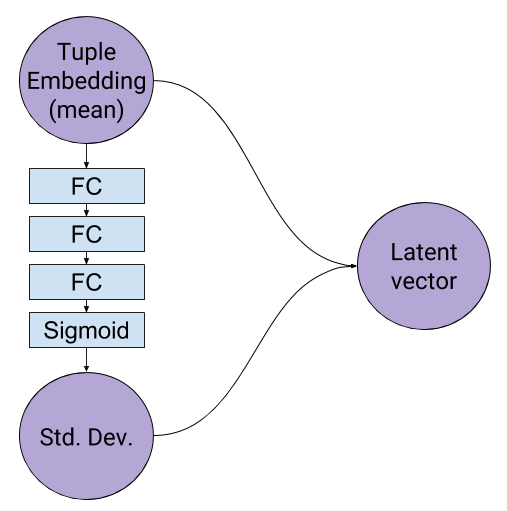}
\label{fig:Standard_Deviation_Network}
\end{figure}

Here we provide the implementation details of the \texttt{StringLiteral} module (Fig~\ref{fig:StringLiteral}), \texttt{ScalarTuple} module (Fig~\ref{fig:ScalarTuple}), \texttt{Tuple} module (Fig~\ref{fig:Tuple}), and the standard deviation network (Fig~\ref{fig:Standard_Deviation_Network}). If not specified otherwise, continuously differentiable exponential linear unit (CELU) with $\alpha = 3$ \citep{DBLP:journals/corr/Barron17a} is the default activation function in our model. It is chosen for its compatibility with the prior distribution of the latent vector since its image covers 99.865\% of the unit Gaussian distribution. Due to its relatively weak nonlinearity, we simply initialize the weights with in-degree scaled unit variance \footnote{truncated at 2 standard deviations. This is tf.variance\_scaling\_initializer(scale=1.0) in TensorFlow.}, i.e. $\mathcal{N}(0, \frac{1}{\sqrt{n_{in}}})$ and zero bias.

\subsection{The \texttt{StringLiteral} module}

The encoder and decoder of the \texttt{StringLiteral} module are character-RNN based on 128-dim gated recurrent unit (GRU) \citep{DBLP:journals/corr/ChoMGBSB14} with 16-dim trainable character embedding initialized with uniform distribution between 0 and 1 and shared between the encoder and decoder. The GRU differs from the original in that it uses CELU with $\alpha = 3$ whose output is capped at 6 in the same way as ReLU6 \citep{krizhevsky2010convolutional} to prevent blow-up, i.e. given 16-dim character embedding $x_t$ at step $t$ and 128-dim RNN state $h_{t-1}$ at step $t-1$, RNN state $h_t$ at step $t$ is given by

\begin{align*}
z_t &= \sigma_g(W_{z} x_t + U_{z} h_{t-1} + b_z) \\
r_t &= \sigma_g(W_{r} x_t + U_{r} h_{t-1} + b_r) \\
c_t &= \sigma_h(W_{h} x_t + U_{h} (r_t \odot h_{t-1}) + b_h) \\
h_t &= z_t  \odot h_{t-1} + (1 - z_t)  \odot c_t
\end{align*}

where $\sigma_g$ is still the sigmoid function but $\sigma_h(x) = \min(\operatorname{CELU}(x, 3), 6)$. $\{W_{r}, U_{r}\}, b_r$ and $\{W_{h}, U_{h}\}, b_h$ are initialized with in-degree scaled unit variance weights and zero bias, but $\{W_{z}, U_{z}\}, b_z$ are initialized with zero weights and unit bias to make sure that the GRU cell isn't too forgetful from the beginning. 

For the encoder, $h_0$ is the zero vector and a $128 \times 128$ fully-connected layer is applied to the final state of the RNN to generate the embedding. For the decoder, $h_0$ is initialized by running another $128 \times 128$ fully-connected layer on the embedding vector and a softmax layer predicts the $t$th character from $h_{t-1}$. We use cross-entropy loss in nat and normalize such that each string field is given total loss weight 1. e.g. the zip code always has 5 digits, so each of them plus the end-of-string token is given loss weight $\frac{1}{6}$.

\subsection{The \texttt{ScalarTuple} module}

The two float fields (lat and long) are collected and modelled jointly by the \texttt{ScalarTuple} module. The module keeps track of the moving mean $\mu$ and the moving covariance matrix $\Sigma$ of the training data, i.e. given values of the scalar tuple in a mini-batch $\mathcal{B} = \{x_{1 \dots m}\}$

\begin{align*}
\mu_{\mathcal{B}} &\gets \frac{1}{m} \sum_{i=1}^{m} x_i \\
\Sigma_{\mathcal{B}} &\gets \operatorname{cov}(\mathcal{B}) \\
\mu &\gets \alpha \mu + (1 - \alpha) \mu_{\mathcal{B}} \\
\Sigma &\gets \alpha \Sigma + (1 - \alpha) \Sigma_{\mathcal{B}}
\end{align*}

where $\alpha$ is the moving average decay, set to be 0.999. The \texttt{ScalarTuple} module then performs PCA-whitening on the raw input for both training and testing:

\begin{align}
UDV^T &= (\Sigma + \epsilon I) \\
x_{whitened} &\gets (x - \mu) U D^{\odot - \frac{1}{2}} \label{eq:whitening}
\end{align}

where $I$ is the identity matrix, $\epsilon$ is a regularization coefficient set to be $10^{-5}$, and $^{\odot - \frac{1}{2}}$ denotes element-wise inverse square root. The encoder then generates the embedding from $x_{whitened}$ with a $2 \times 128$ sigmoid layer, and the decoder generates $x'_{whitened}$ from the embedding with a linear $128 \times 2$ layer and computes squared error loss $\sum (x_{whitened} - x'_{whitened})^2$. Both float fields are given total loss weight 1, so sum of the squared error is used instead of average. PCA-whitening is similar to and reduces to batch normalization \citep{DBLP:journals/corr/IoffeS15} without scale and shift when components of $x$ is uncorrelated, but automatically handles strong correlation which can cause batch normalization to overestimate the true variances of the data. For generation, the inverse of Eq \eqref{eq:whitening} is used to un-whiten the prediction.

\subsection{The \texttt{Tuple} module}

Unlike the \texttt{StringLiteral} module and the \texttt{ScalarTuple} module, the \texttt{Tuple} module is not a leaf module, i.e. the input of its encoder is the embeddings generated by the encoders of its child modules, and the output of its decoder is the embeddings used by the decoders of its child modules. The encoder is based on a bidirectional RNN with the same GRU implementation as the \texttt{StringLiteral} module, but with 128-dim state size and 128-dim input size. In addition, since size of the tuple is fixed and each element of the tuple is different, GRU cells for different tuple elements are distinct and do not share parameters. For example, with 7 string fields and 2 float fields modelled by a \texttt{ScalarTuple} module, the \texttt{Tuple} module encoder for the address model has $8 \times 2 = 16$ distinct GRU cells. The bidirectional RNN has a shared 128-dim trainable initial state for both directions, and the two final states of the bidirectional RNN are concatenated and fed to a $256 \times 128$ fully-connected layer to produce the final embedding.

The \texttt{Tuple} module decoder is also based on an RNN. The initial state for the decoder RNN is initialized from the embedding with a $128 \times 128$ fully-connected layer. Each element of the tuple again has its own GRU cell, and the same GRU implementation is used with  128-dim state size and 128-dim input size. In addition, each GRU cell includes a $128 \times 128$ fully-connected layer that generates its child module embedding from the current state. This child module embedding is then fed back to the GRU cell to increment to the next state for generation and scheduled sampling \citep{DBLP:journals/corr/BengioVJS15}. For training/testing with teacher forcing, embedding given by the child encoder is used as the ground-truth input. In order to make sure that the encoder and decoder of the child module use the same representation, we add an extra loss term dubbed skew loss, which is the mean squared error between the generated embedding and the embedding given by the child encoder. This skew loss is somewhat arbitrarily given the same weight as the reconstruction loss of the respective child module.

The model described here is generated from the Protocol Buffer message definition by an internal framework. Code-named Alala in reference to the Greek goddess and the Hawaiian crow \textit{Corvus hawaiiensis}, the framework is based on TensorFlow Fold \citep{DBLP:journals/corr/LooksHHN17} and developed for training VAEs on arbitrarily-defined protocol buffers. The order of the elements of the tuple follows that of the Protocol Buffer message definition, except that (lat, long) are collected and modelled jointly by the \texttt{ScalarTuple} module as the last element of the tuple.

\subsection{Standard deviation network}
\label{sec:standard_deviation_network}

Encoders of Alala modules produce an embedding vector. In order to train a VAE with Alala modules, we interpret the embedding vector as the mean vector $\mu$ and generate the standard deviation vector $\sigma(\mu)$ from it with a standard deviation network. For the model described here, the standard deviation network consists of 3 $128 \times 128$ fully-connected layers, topped with a sigmoid layer to produce the standard deviation vector whose elements are always in the range $(0, 1)$. The sigmoid layer is initialized with zero weights and -5 bias to make sure that the standard deviation vector starts out small in the beginning of the training process.

\clearpage

\section{Full hyperparameter sweep result}
\label{appendix:hyperparameter_sweep}

The $\beta$-VAE baseline experiments use the full 2M steps as the warm-up period, and the ground-truth probability decreases linearly from 1 to 0 for scheduled sampling experiments. For a rough measure of the reproducibility, we rerun the best experiments with the same hyperparameters.

\begin{table}[ht]
  \caption{$\beta$-VAE baseline performance}\label{tab:baseline}
  \begin{adjustbox}{center}
  \begin{tabular}{| c | c | c | c | c | c | c |}
    \hline
    & $\beta_{\mathrm{start}}$ & $\beta_{\mathrm{end}}$ & mean & median & stddev & $\bar{d}_{\mathrm{Levenshtein}}$ \\ %
    \hline \hline
    \multirow{2}{*}{Tuple SS + String TF} & 0 & 0.384 & 0.246 -- \textbf{0.261} & 0.0450 -- \textbf{0.0606} & 0.321 -- 0.329 & 0.114 -- 0.111 \\ %
    \cline{2-7}
    & \multicolumn{2}{|c|}{0.128} & 0.137 & < 0.01 & 0.267 & 0.600 \\ %
    \hline
    \multirow{2}{*}{Tuple AS + String TF} & 0 & 0.384 & 0.240 -- 0.249 & 0.0448 -- 0.0505 & 0.317 -- 0.322 & 0.184 -- 0.149 \\ %
    \cline{2-7}
    & \multicolumn{2}{|c|}{0.128} & 0.179 & < 0.01 & 0.298 & 0.487 \\ %
    \hline
    \multirow{2}{*}{Always Sampling} & 0 & 0.384 & 0.173 & < 0.01 & 0.294 & 0.0306 \\ %
    \cline{2-7}
    & \multicolumn{2}{|c|}{0.128} & 0.179 -- 0.203 & < 0.01 & 0.293 -- 0.303 & 0.0234 -- 0.0511 \\ %
    \hline
    \multirow{7}{*}{Scheduled Sampling} & \multirow{3}{*}{0} & 0.64 & 0.225 & < 0.01 & 0.326 & 0.144 \\ %
    \cline{3-7}
    & & 0.384 & 0.215 -- 0.247 & < 0.01 -- 0.0237 & 0.317 -- 0.331 & 0.0865 -- 0.0974 \\ %
    \cline{3-7}
    & & 0.128 & 0.208 & < 0.01 & 0.309 & 0.0445 \\ %
    \cline{2-7}
    & \multicolumn{2}{|c|}{0.64} & 0.0864 & 0 & 0.220 & 0.884 \\ %
    \cline{2-7}
    & \multicolumn{2}{|c|}{0.384} & 0.103 & 0 & 0.240 & 0.751 \\ %
    \cline{2-7}
    & \multicolumn{2}{|c|}{0.128} & 0.119 -- 0.248 & < 0.01 -- 0.0361 & 0.253 -- 0.325 & 0.291 -- 0.166 \\ %
    \cline{2-7}
    & \multicolumn{2}{|c|}{0.064} & < 0.01 & 0 & 0.00365 & 0.975 \\ %
    \hline
    \multirow{2}{*}{Teacher Forcing} & 0 & 0.384 & 0.0178 & 0 & 0.0970 & 0.0961 \\ %
    \cline{2-7}
    & \multicolumn{2}{|c|}{0.128} & < 0.01 & 0 & 0.0283 & 0.535 \\ %
    \hline
  \end{tabular}
  \end{adjustbox}
\end{table}

For augmented training experiments below, $p_{\mathrm{sampled}} = \nicefrac{1}{5}$, \texttt{gen\_start\_step} = $2 \times 10^4$ for scheduled sampling (roughly when generated loss stops decreasing) and \texttt{gen\_start\_step} = $2 \times 10^5$ for Tuple SS + String TF (roughly when training loss stops rapidly decreasing). We use $n_{\mathrm{augmented}}$ = 256 augmented latent vectors, so training batches now consist of 256 training examples and 256 generated variants. In practice, data generation is slow due to the lack of parallelism, so we actually shut down the initial training process with 32 workers after \texttt{gen\_start\_step} and relaunch it with 512 workers for augmented training. These augmented training experiments always employ simultaneous scheduled sampling and KL-divergence warm-up, with $\beta_{\mathrm{start}} = 0$ and the first 1M training steps as the warm-up period. We also find that it's beneficial to have a KL-divergence cool-down period after the warm-up period, in which case we have $\beta_{\mathrm{mid}} > \beta_{\mathrm{end}}$. %

\begin{table}[ht]
\centering
  \caption{Augmented training performance}\label{tab:augmented}
  \begin{tabular}{| c | c | c | c | c | c | c |}
    \hline
    $p_{\mathrm{sampled}}$ & $\beta_{\mathrm{mid}}$ & $\beta_{\mathrm{end}}$ & mean & median & stddev & $\bar{d}_{\mathrm{Levenshtein}}$ \\ %
    \hline
    \multicolumn{7}{|c|}{Tuple SS + String TF} \\
    \hline
    \multirow{2}{*}{$\nicefrac{1}{5}$} & \multicolumn{2}{|c|}{0.128} & 0.316 & 0.123 & 0.356 & 0.0402 \\ %
    \cline{2-7}
    & 0.64 & 0.128 & 0.401 -- \textbf{0.401} & 0.321 -- \textbf{0.324} & 0.373 -- 0.376 & 0.239 -- 0.195 \\ %
    \hline
    \multicolumn{7}{|c|}{Scheduled Sampling} \\
    \hline
    $\nicefrac{1}{5}$ & \multicolumn{2}{|c|}{\multirow{3}{*}{0.128}} & 0.272 & 0.0327 & 0.348 & 0.0266 \\ %
    \cline{1-1} \cline{4-7}
    $\nicefrac{1}{3}$ & \multicolumn{2}{|c|}{} & 0.279 & 0.0481 & 0.349 & 0.0207 \\ %
    \cline{1-1} \cline{4-7}
    $\nicefrac{1}{2}$ & \multicolumn{2}{|c|}{} & 0.275 & 0.0348 & 0.350 & 0.0242 \\ %
    \hline
    $\nicefrac{1}{8}$ & \multirow{5}{*}{0.64} & \multirow{5}{*}{0.128} & 0.311 -- 0.333 & 0.105 -- 0.159 & 0.360 -- 0.362 & 0.113 -- 0.112 \\ %
    \cline{1-1} \cline{4-7}
    $\nicefrac{1}{5}$ & & & 0.317 -- 0.335 & 0.137 -- 0.180 & 0.354 -- 0.358 & 0.116 -- 0.115 \\ %
    \cline{1-1} \cline{4-7}
    $\nicefrac{1}{3}$ & & & 0.307 & 0.106 & 0.355 & 0.115 \\ %
    \cline{1-1} \cline{4-7}
    $\nicefrac{1}{2}$ & & & 0.312 & 0.110 & 0.359 & 0.114 \\ %
    \cline{1-1} \cline{4-7}
    1 & & & 0.311 & 0.110 & 0.358 & 0.108 \\ %
    \hline
  \end{tabular}
\end{table}

For multiscale VAE experiments below, no augmented training is used and they always employ scheduled sampling with the first 1M training steps as the warm-up period. In case KL-divergence warm-up or cool-down is also employed during the warm-up period, we have $\beta_{\mathrm{max, start}} \neq \beta_{\mathrm{max, end}}$. For experiments combining this setup and augmented training, see Appendix \ref{appendix:multiscale_augmented}.

\begin{table}[ht]
\centering
  \caption{Multiscale VAE only performance}\label{tab:multiscale}
  \begin{tabular}{| c | c | c | c | c | c |}
    \hline
    $\beta_{\mathrm{max, start}}$ & $\beta_{\mathrm{max, end}}$ & mean & median & stddev & $\bar{d}_{\mathrm{Levenshtein}}$ \\ %
    \hline
    \multicolumn{6}{|c|}{Tuple SS + String TF} \\
    \hline
    1.28 & 0.64 & 0.476 -- \textbf{0.509} & \textbf{0.494} -- 0.580 & 0.382 -- 0.384 & 0.916 -- 0.921 \\ %
    \hline
    \multicolumn{2}{|c|}{1.28} & 0.460 & 0.474 & 0.373 & 0.921 \\ %
    \hline
    \multicolumn{6}{|c|}{Scheduled Sampling} \\
    \hline
    5.12 & \multirow{4}{*}{0.64} & 0.386 & 0.297 & 0.368 & 0.605 \\ %
    \cline{1-1} \cline{3-6}
    2.56 & & 0.376 & 0.275 & 0.367 & 0.619 \\ %
    \cline{1-1} \cline{3-6}
    1.28 & & 0.402 -- 0.411 & 0.349 -- 0.361 & 0.367 -- 0.371 & 0.497 -- 0.525 \\ %
    \cline{1-1} \cline{3-6}
    0 & & 0.218 & 0.00738 & 0.315 & 0.0441 \\ %
    \hline
    \multicolumn{2}{|c|}{5.12} & 0.224 & 0.00101 & 0.343 & 0.965 \\ %
    \hline
    \multicolumn{2}{|c|}{2.56} & 0.396 & 0.306 & 0.379 & 0.685 \\ %
    \hline
    \multicolumn{2}{|c|}{1.28} & 0.395 & 0.326 & 0.367 & 0.581 \\ %
    \hline
    \multicolumn{2}{|c|}{0.64} & 0.313 & 0.152 & 0.344 & 0.268 \\ %
    \hline
  \end{tabular}
\end{table}

\clearpage

\section{Map interpolations}
\label{appendix:map_interpolatios}

Map interpolations of the models featured in the main text. For the live version and more interpolation examples, see the HTML files of the repository \url{https://github.com/EIFY/vermont\_address}.

\begin{figure}[ht]
\centering
\caption{Interpolation between the first 2 training examples by the VAE trained with KL-divergence warm-up (Tuple SS + String TF).}\label{fig:warm_up_interpolation}
\includegraphics[width=\textwidth]{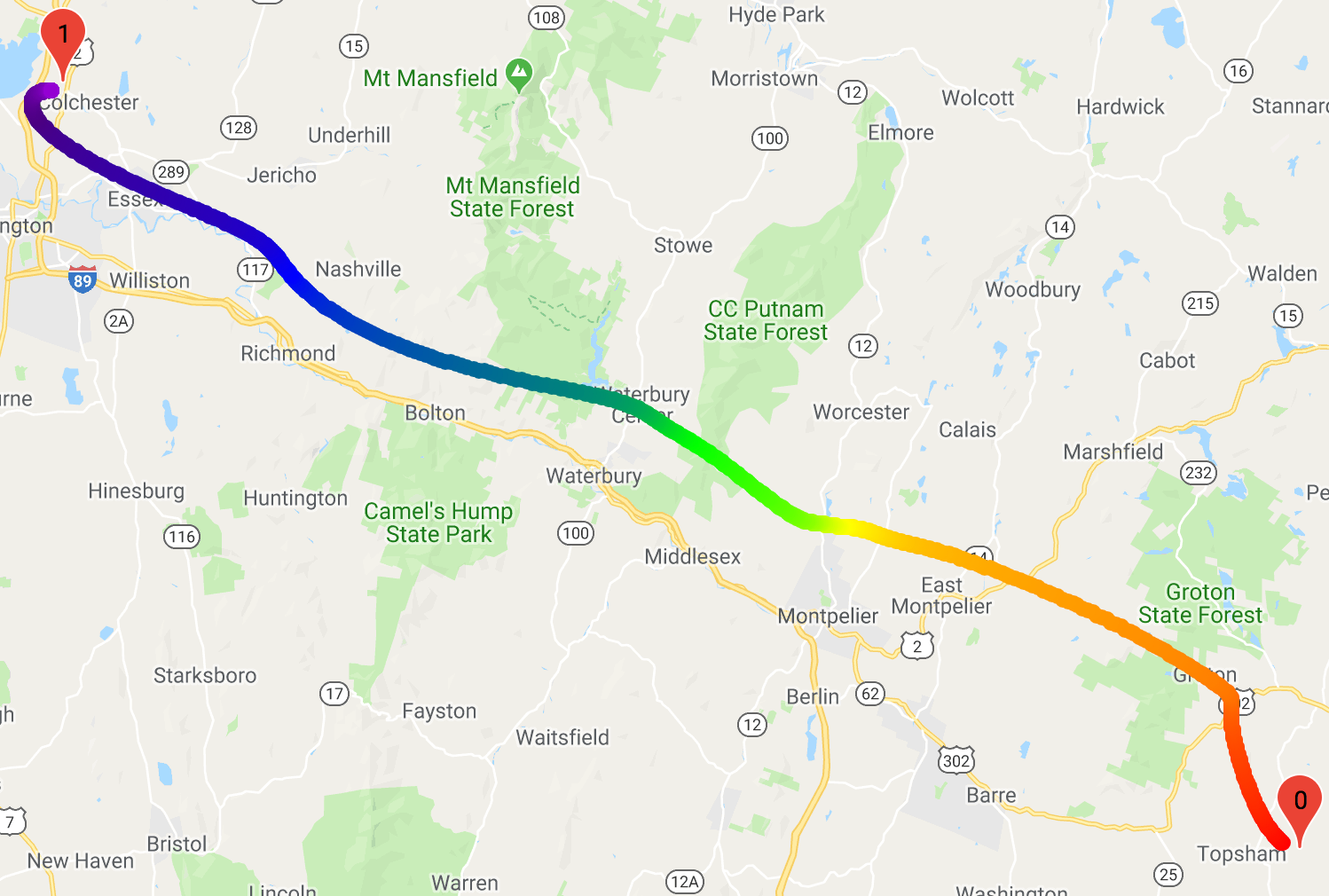}
\end{figure}

\begin{figure}[ht]
\centering
\caption{Interpolation between the first 2 training examples by the VAE trained with augmented training.}\label{fig:augmented_interpolation}
\includegraphics[width=\textwidth]{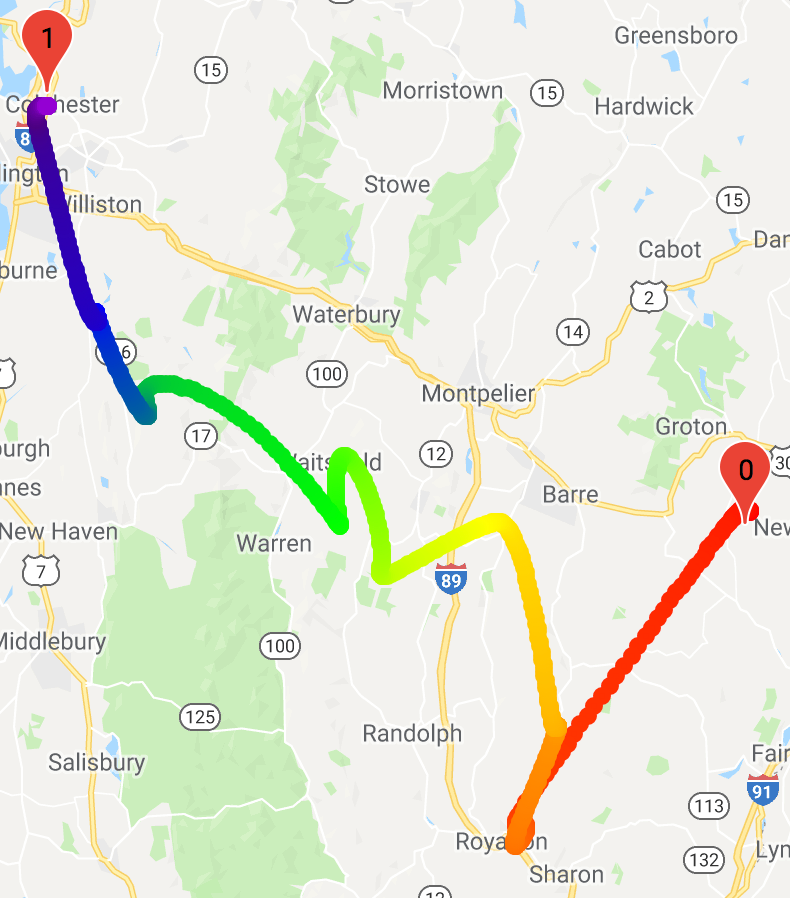}
\end{figure}

\begin{figure}[ht]
\centering
\caption{Interpolation between the first 2 training examples by the multiscale VAE (without augmented training).}\label{fig:multiscale_interpolation}
\includegraphics[width=\textwidth]{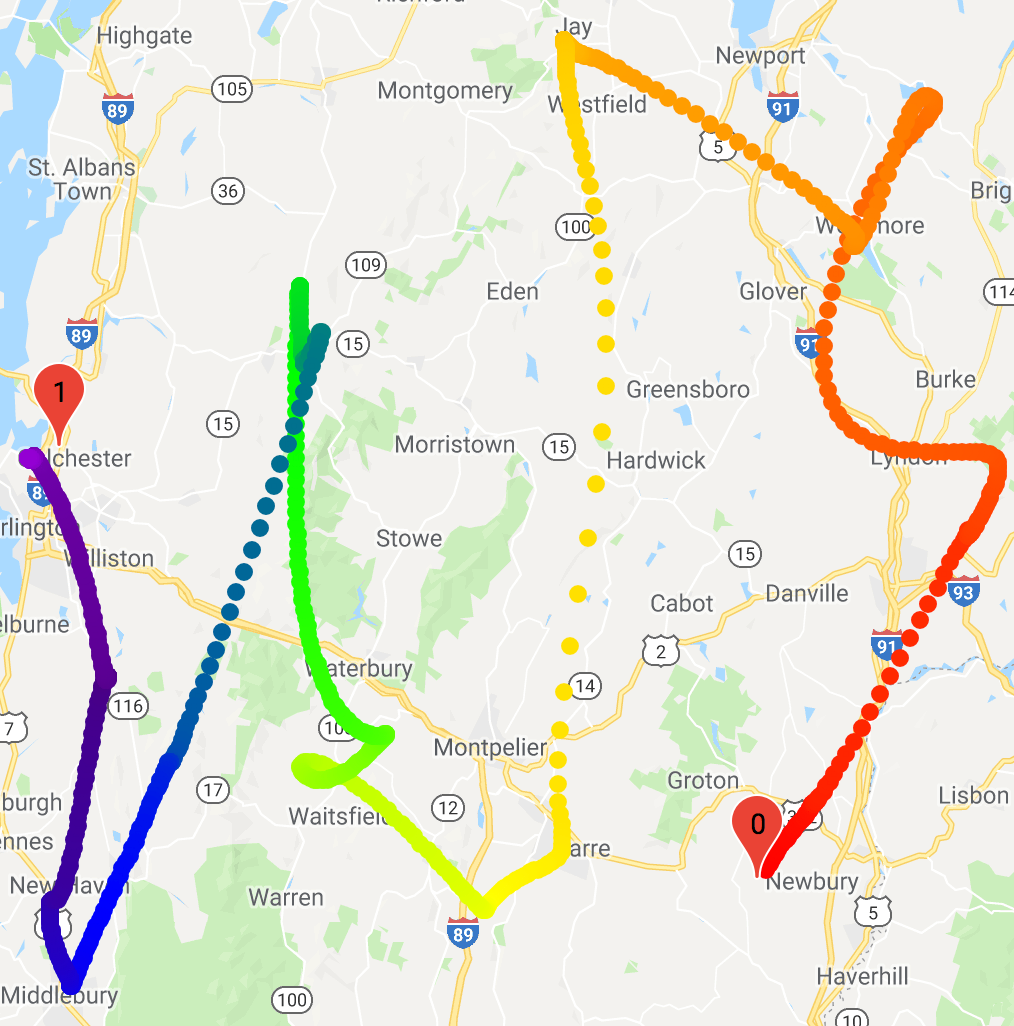}
\end{figure}

\clearpage

\section{Stats over repeated encoding and decoding}
\label{appendix:repeated_encoding_decoding}

In the following, we plot the p-value distributions of 10000 generated samples and 10000 training examples over repeated encoding and decoding for the models featured in the main text. For generated samples $\tilde{x}$, we examine the p-values of the following sequence:

\begin{align*}
\tilde{x} &= g(z) \\
\tilde{x}' &= g(\mu_\lambda(\tilde{x})) \\
\tilde{x}'' &= g(\mu_\lambda(\tilde{x}')) \\
&\dots \\
\tilde{x}^{(n)} &= g(\mu_\lambda(\tilde{x}^{(n-1)})) \text{ for } n > 0\\
\end{align*}

For training examples $x$, we examine the p-values of the following sequence:

\begin{align*}
x' &= g(\mu_\lambda(x)) \\
x'' &= g(\mu_\lambda(x')) \\
&\dots \\
x^{(n)} &= g(\mu_\lambda(x^{(n-1)})) \text{ for } n > 0 \\
\end{align*}

For $n$ from 0 to 9. The results strongly suggest that $x^{(n)} \overset{d}{=} x'$ for $n > 1$ and $\lim_{n\to\infty} \tilde{x}^{(n)} \overset{d}{=} x'$ where $\overset{d}{=}$ indicates that two random variables follow the same distribution.

\begin{figure}[ht]
\centering
\caption{Box plot of p-values over repeated encoding and decoding (Tuple SS + String TF). The `generated' sequence at repetition = $n$ corresponds to the p-value distribution of 10000 samples of $\tilde{x}^{(n)}$, and the `reconstructed' sequence at repetition = $n$ corresponds to that of 10000 randomly selected training examples $x^{(n)}$.}\label{fig:warm_up_teacher_str_repeated_boxes}
\includegraphics[width=\textwidth]{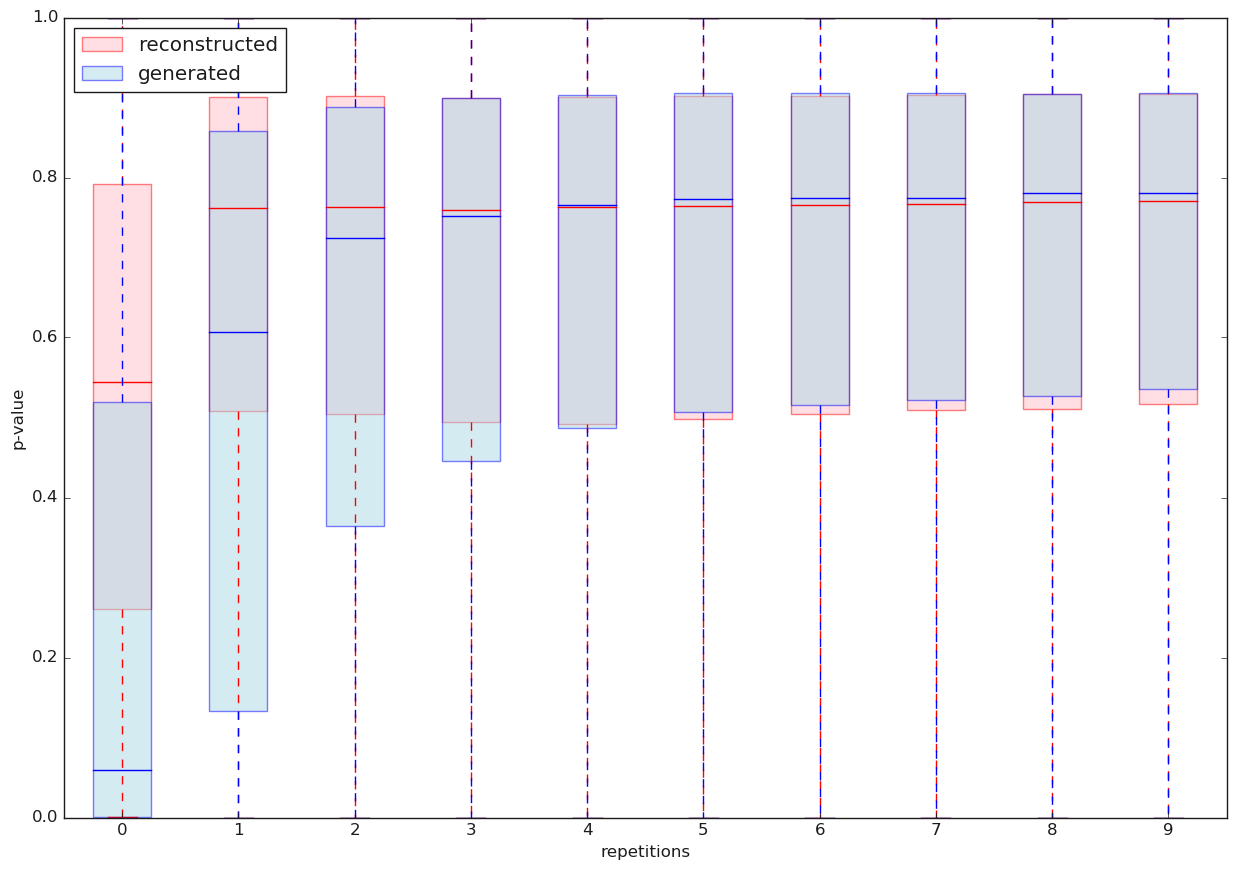}
\end{figure}

\begin{figure}[ht]
\centering
\caption{Box plot of p-values over repeated encoding and decoding (augmented training). Some of the p-values are unaffected by repeated encoding and decoding and show up as outliers more than 1.5 IQR (interquartile range) away from the lower quartile.}\label{fig:augmented_repeated_boxes}
\includegraphics[width=\textwidth]{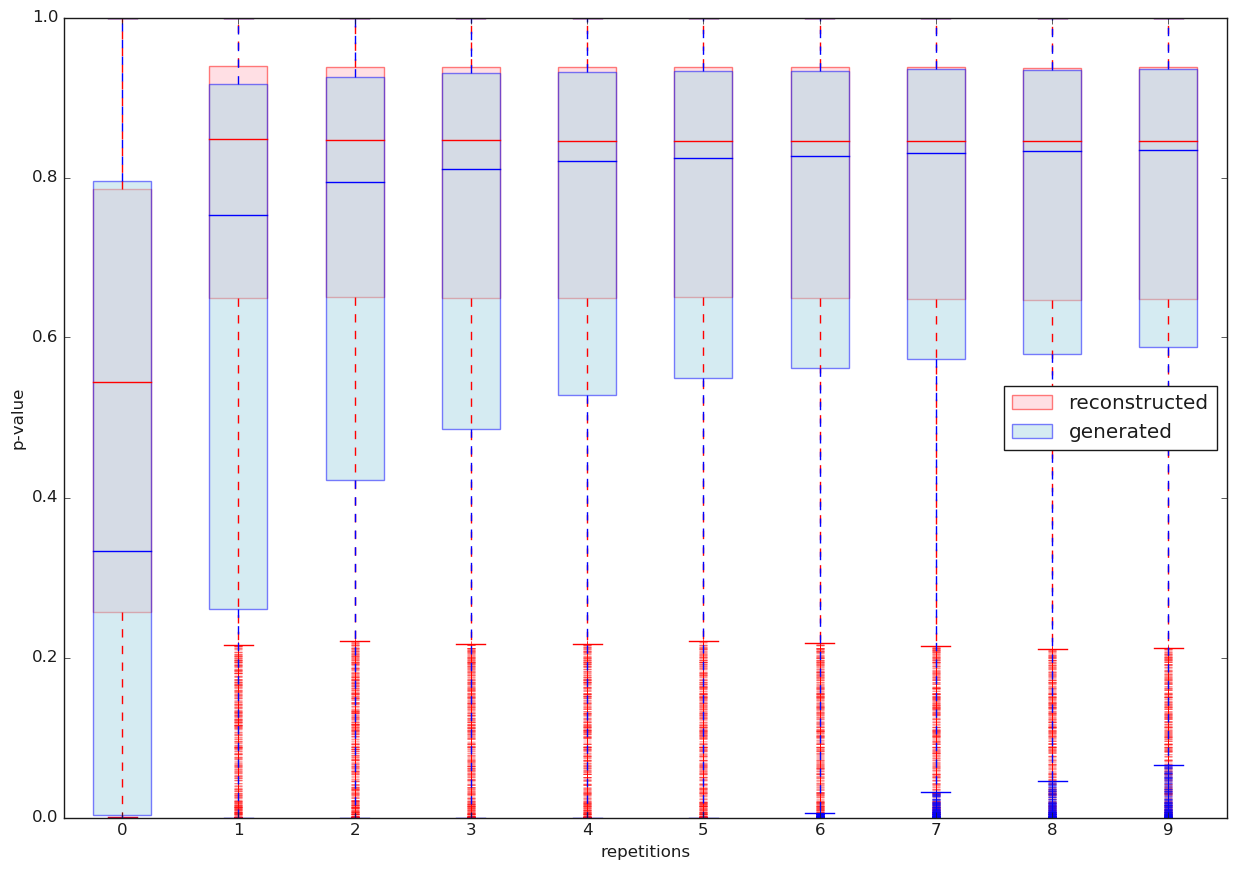}
\end{figure}

\begin{figure}[ht]
\centering
\caption{Box plot of p-values over repeated encoding and decoding (multiscale VAE only)}\label{fig:multiscale_repeated_boxes}
\includegraphics[width=\textwidth]{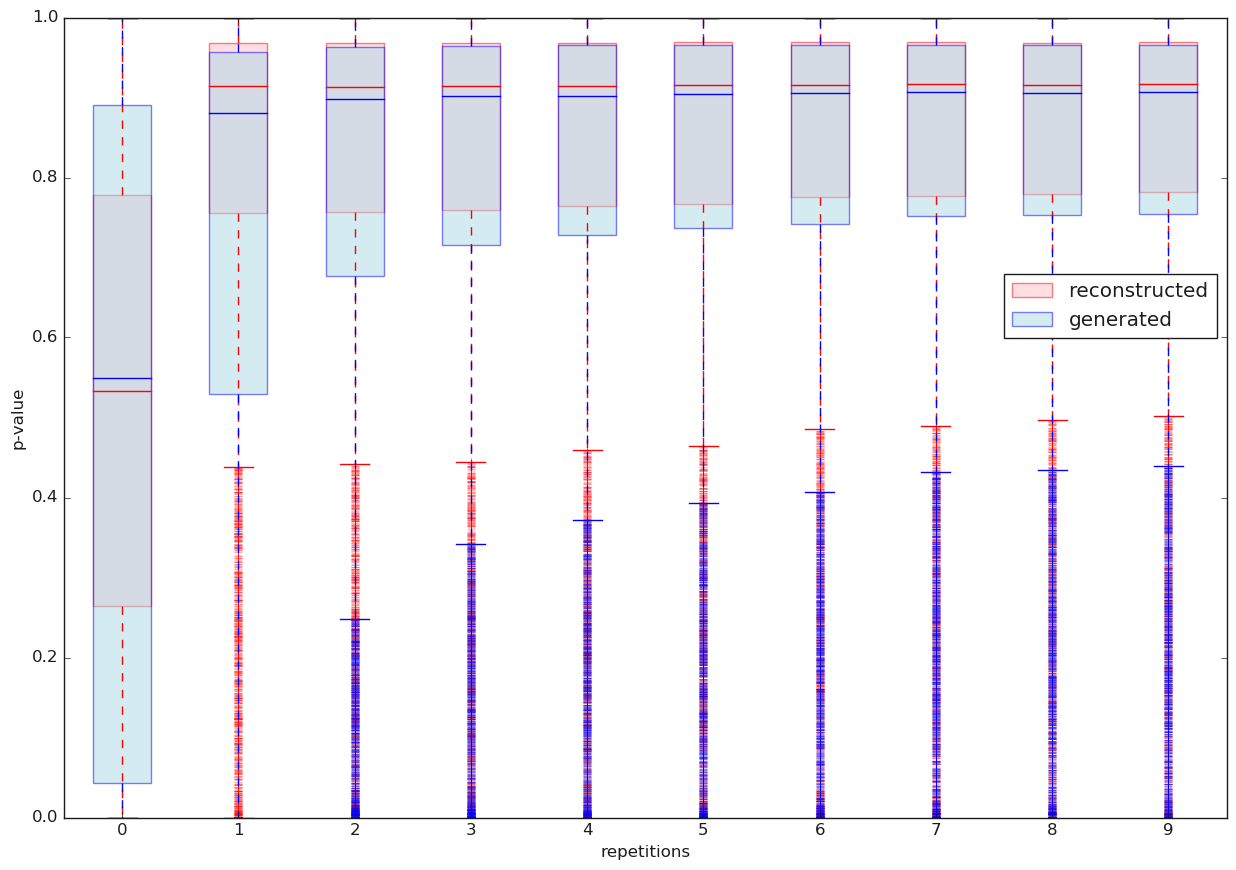}
\end{figure}

\clearpage

We also examine the proportion of street names of the generated samples present in the training data over repeated encoding and decoding. For models that encode the street names, the proportion also increases over repeated encoding and decoding. This observation has privacy implications: Namely, repeated encoding and decoding may be an efficient tool to extract rare or unique sequence of the training data from a VAE. It is instrumental to quantify such risk a la \citet{DBLP:journals/corr/abs-1802-08232}, and we leave it to future work.

\begin{figure}[ht]
\centering
\caption{Number of generated street names present in the training data out of 10000 samples of $\tilde{x}^{(n)}$ over repeated encoding and decoding (Tuple SS + String TF)}\label{fig:warm_up_teacher_str_street_names}
\includegraphics[width=\textwidth]{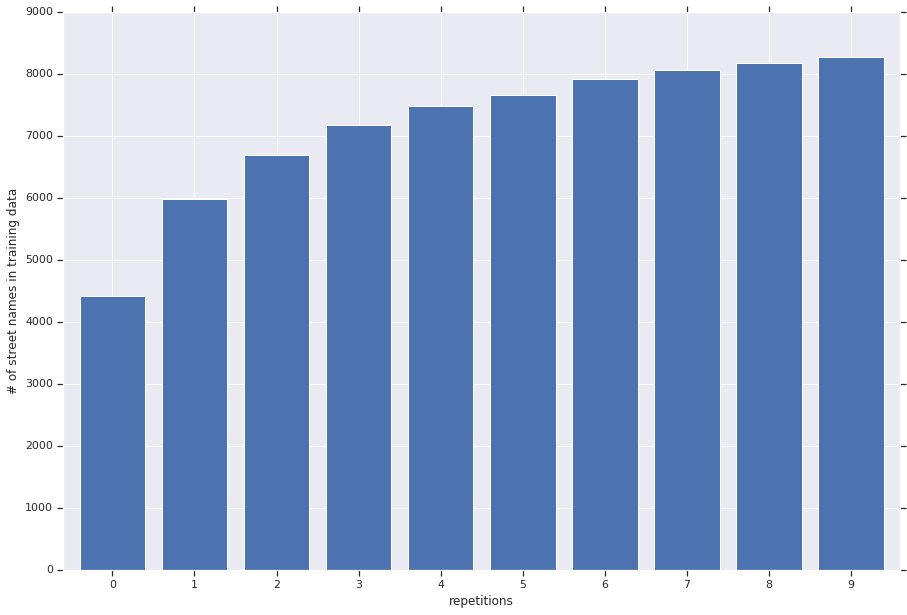}
\end{figure}

\begin{figure}[ht]
\centering
\caption{Number of generated street names present in the training data out of 10000 samples of $\tilde{x}^{(n)}$ over repeated encoding and decoding (augmented training).}\label{fig:augmented_teacher_str_street_names}
\includegraphics[width=\textwidth]{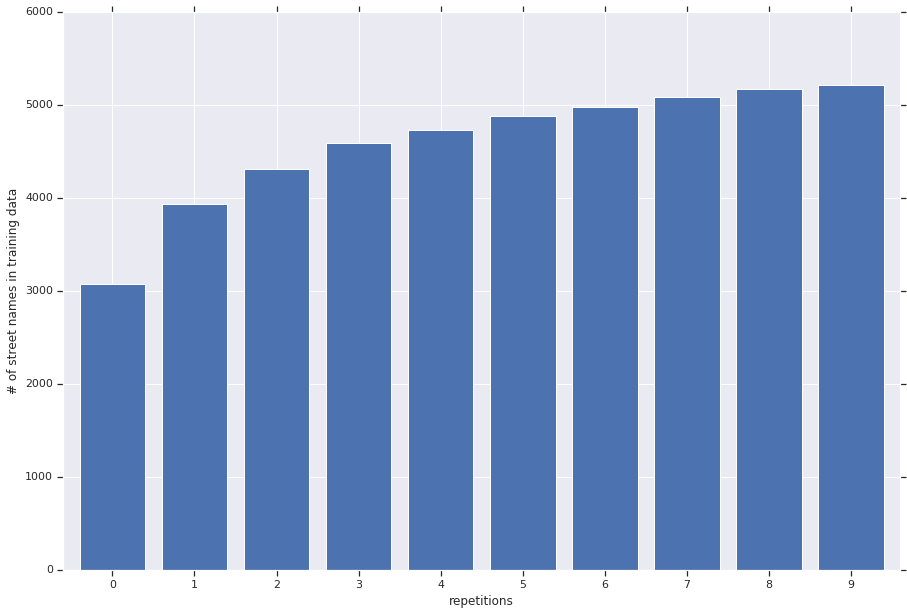}
\end{figure}

\begin{figure}[ht]
\centering
\caption{Number of generated street names present in the training data out of 10000 samples of $\tilde{x}^{(n)}$ over repeated encoding and decoding (multiscale VAE only). The proportion of generated street names present in the training data is higher than that of optimized $\beta$-VAE (Fig~\ref{fig:warm_up_teacher_str_street_names}) and stays constant throughout repeated encoding/decoding.}\label{fig:multiscale_teacher_str_street_names}
\includegraphics[width=\textwidth]{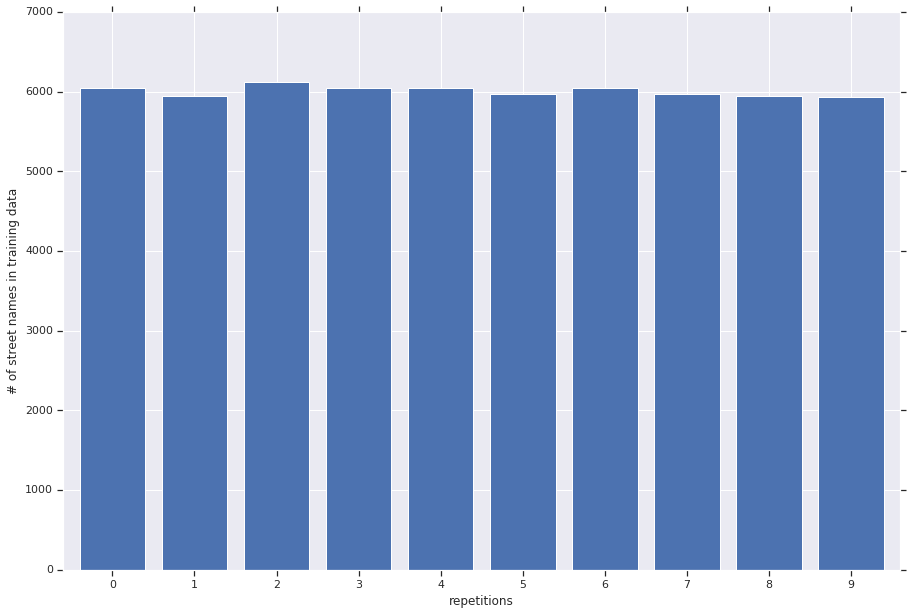}
\end{figure}

\clearpage

\section{Comma-separated text model}
\label{appendix:text_model}

The simplest approach to model the Vermont state address data is to model them in their original format as comma-separated text. So we trained a simple seq-to-seq model using just the \texttt{StringLiteral} module and the standard deviation network  (Table~\ref{tab:comma-separated}). We doubled the size of the embedding vector to 256-dim to compensate for the difference in number of parameters, omitted the string fields that are always empty so we only expect 6 comma-separated values (number, street, city, postcode, lat, long), and rounded the floating point numbers for the coordinates to five decimal places. If the generated comma-separated text does not have enough comma-separated values, or the last two values do not represent valid input for python's \texttt{float()} function, we consider the generated sample to be malformed and its p-value to be zero. We used teacher forcing as the training scheme for these experiment as scheduled sampling simply does not work. $\beta_{\mathrm{start}} = 0, \beta_{\mathrm{end}} = 0.768$ for the KL divergence warm-up experiment, but running multiscale VAE with $\beta_{\mathrm{max, start}} = 2.56$ and $\beta_{\mathrm{max, end}} = 1.28$ resulted in posterior collapse. $\beta_{\mathrm{max}}$ is fixed at 0.768 for the other multiscale VAE experiment.

\begin{table}[ht]
\centering
  \caption{Comma-separated text model performance}\label{tab:comma-separated}
  \begin{tabular}{| c | c | c | c | c |}
    \hline
    & malformed & mean & median & stddev \\ %
    \hline \hline
    KL warm-up & 23 & 0.424 & 0.411 & 0.344 \\ %
    \hline
    Posterior collapse & \textbf{17} & \textbf{0.436} & \textbf{0.428} & 0.340 \\ %
    \hline
    Multiscale VAE & 43 & 0.394 & 0.356 & 0.344 \\ %
    \hline
  \end{tabular}
\end{table}

\begin{figure}[ht]
\centering
\caption{Interpolation between the first 2 training examples by the comma-separated text model (KL warm-up).}\label{fig:string_cat_teacher_interpolation}
\includegraphics[width=\textwidth]{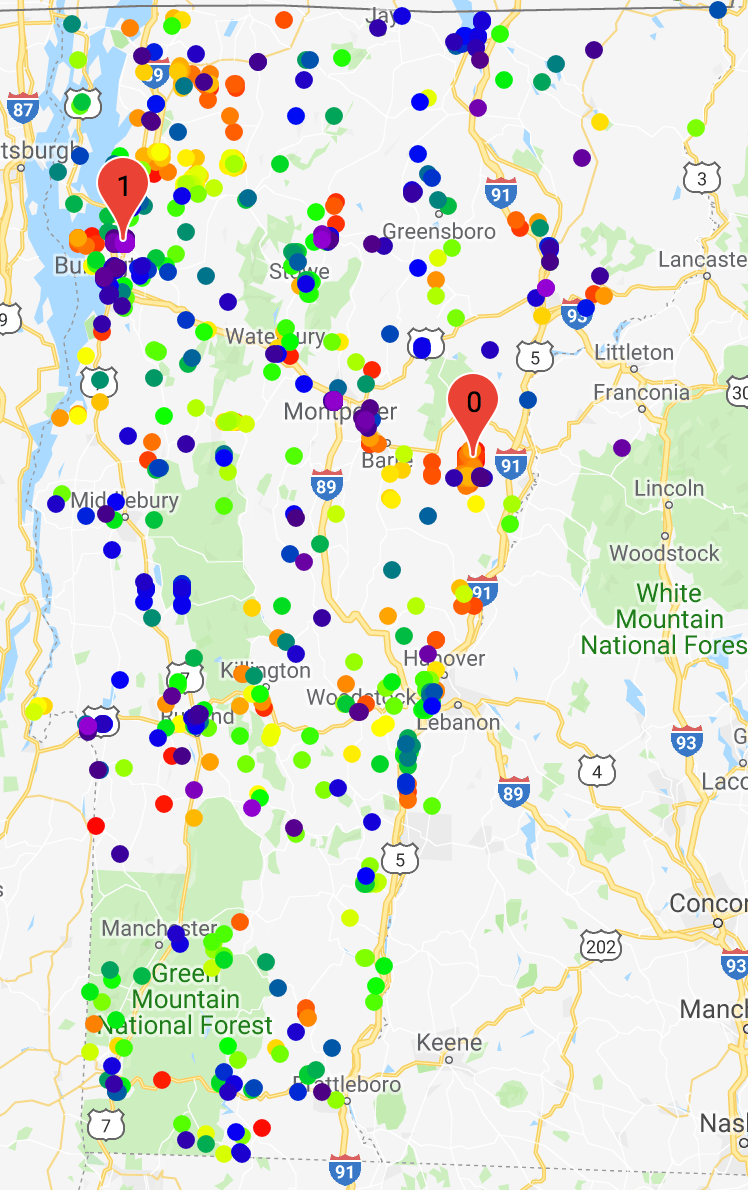}
\end{figure}

\begin{figure}[ht]
\centering
\caption{"Interpolation" between the first 2 training examples by the comma-separated text model (posterior collapse). More purple and red-to-orange markers are visible here in comparison to Fig~\ref{fig:string_cat_teacher_interpolation} since they are no longer concentrated around their closest training examples.}\label{fig:string_cat_teacher_collapsed_interpolation}
\includegraphics[width=\textwidth]{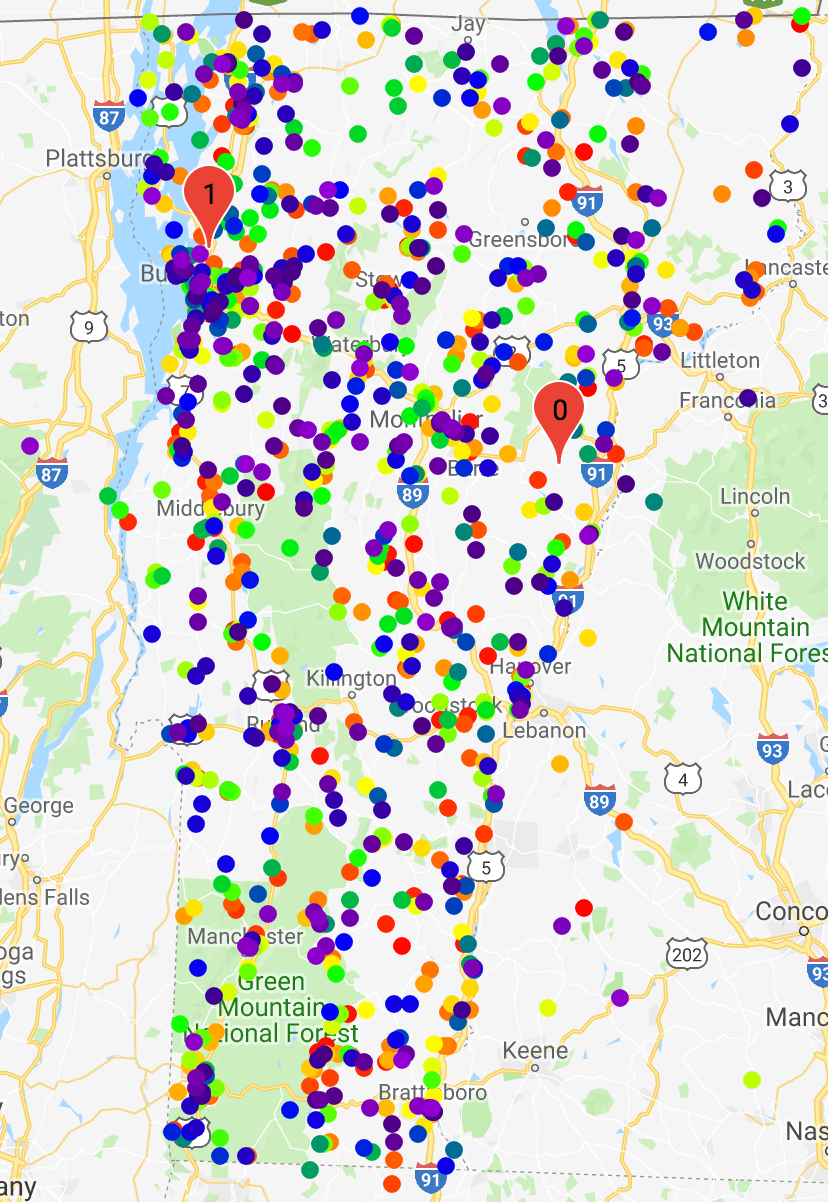}
\end{figure}

We can see that these comma-separated text models are quite good at generating valid samples (with fewer than 0.5\% of 10000 samples malformed) and capturing zip-coordinate correlations. While it is possible to train a meaningful latent comma-separated text model (Fig~\ref{fig:string_cat_teacher_interpolation}), it does not improve the generation quality over a pure autoregressive model resulted from posterior collapse (Fig~\ref{fig:string_cat_teacher_collapsed_interpolation}). We speculate that per-character cross-entropy reconstruction loss function simply does not yield a latent space with good structure -- from the model's perspective addresses that start with "147,HARTS RD" are the addresses closest to each other, not addresses that are geographically close.

\clearpage

\section{Pass-through model}
\label{appendix:pass_through_model}

To establish a baseline against the tree recursive model, we implemented a simple pass-through model by replacing the \texttt{Tuple} module with a \texttt{SimpleTuple} module. The decoder of the \texttt{SimpleTuple} module just passes the embedding vector down to child decoders, and the encoder just concatenates embedding vectors generated by child encoders and applies a fully-connected to generate the final embedding. This architecture requires separate string models for each of the four non-empty string fields (number, street, city, and postcode) and we just omitted the string fields that are always empty. We then used the same p-value metric to quantify the generation quality of pass-through models in the table below. Other than $p_{\mathrm{sampled}}=0.2$ and \texttt{gen\_start\_step} = $8 \times 10^4$ for augmented training, hyperparameters not specified in the table are the same as in the corresponding tree recursive model experiments.

\begin{table}[htbp]
\centering
  \caption{Pass-through model performance}\label{tab:pass-through}
  \begin{tabular}{| c | c | c | c | c | c |}
    \hline
    & $\beta_{\mathrm{start}}$ & $\beta_{\mathrm{end}}$ & mean & median & stddev \\ %
    \hline
    \multicolumn{6}{|c|}{Generalized $\beta$-VAE} \\
    \hline
    Always Sampling & \multicolumn{2}{|c|}{\multirow{2}{*}{0.128}} & 0.0699 & $9.53 \times 10^{-7}$ & 0.176 \\ %
    \cline{1-1} \cline{4-6}
    \multirow{2}{*}{Scheduled Sampling} & \multicolumn{2}{|c|}{} & 0.0603 & $1.17 \times 10^{-7}$ & 0.168 \\ %
    \cline{2-6}
    & \multirow{2}{*}{0} & \multirow{2}{*}{0.384} & 0.0652 & $2.68 \times 10^{-7}$ & 0.175 \\ %
    \cline{1-1} \cline{4-6}
    Teacher Forcing & & & \textbf{0.0835} & $\mathbf{1.33 \times 10^{-4}}$ & 0.192 \\ %
    \hline
    \multicolumn{6}{|c|}{Augmented Training} \\
    \hline
    Scheduled Sampling & \multicolumn{2}{|c|}{\multirow{2}{*}{$0 \to 0.64 \to 0.128$}} & 0.0614 & $5.11 \times 10^{-8}$ & 0.171 \\ %
    \cline{1-1} \cline{4-6}
    Teacher Forcing & \multicolumn{2}{|c|}{} & \textbf{0.118} & \textbf{0.00110} & 0.229 \\ %
    \hline
    \multicolumn{6}{|c|}{Multiscale VAE only} \\
    \hline
    Scheduled Sampling & \multirow{2}{*}{1.28} & \multirow{2}{*}{0.64} & 0.0612 & $1.27 \times 10^{-7}$ & 0.168 \\ %
    \cline{1-1} \cline{4-6}
    Teacher Forcing & & & \textbf{0.0766} & $\mathbf{6.81 \times 10^{-5}}$ & 0.186 \\ %
    \hline
  \end{tabular}
\end{table}

\begin{figure}[htbp]
\centering
\caption{Loss (left) and BPC (right) for training data, testing data, and generated samples during the training process of the pass-through model with teacher forcing.}\label{fig:simple_teacher_generated_loss}
\includegraphics[width=\textwidth]{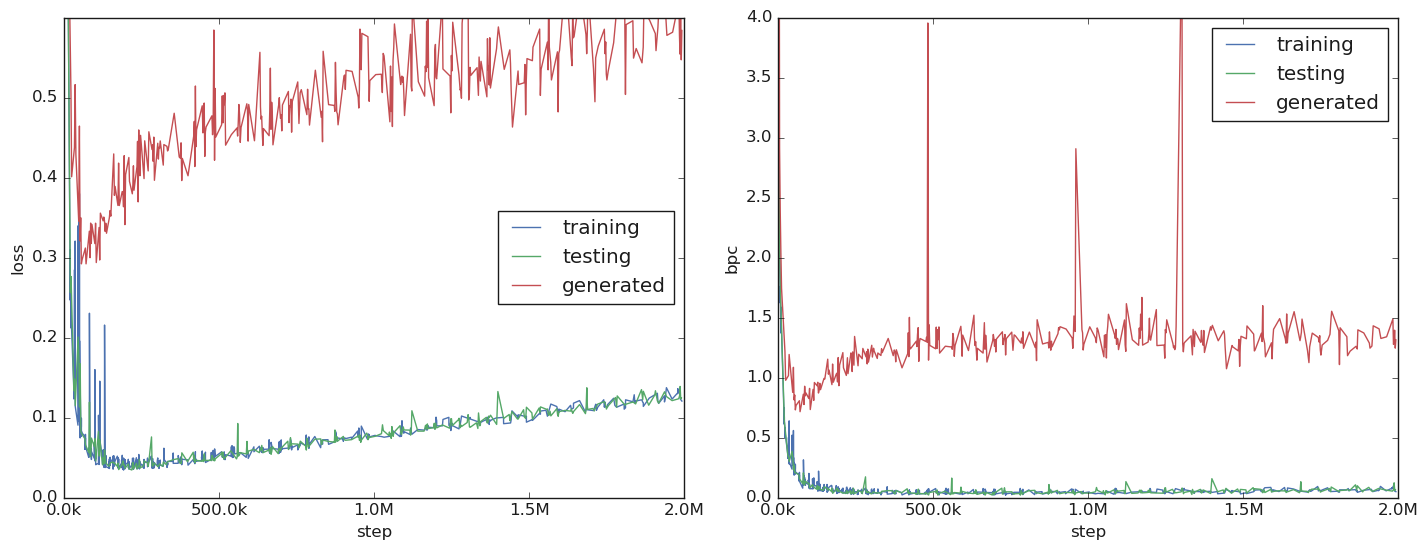}
\end{figure}

\begin{figure}[htbp]
\centering
\caption{Loss (left) and BPC (right) for training data, testing data, and generated samples during the training process of the pass-through model with augmented training and teacher forcing.}\label{fig:simple_augmented_generated_loss}
\includegraphics[width=\textwidth]{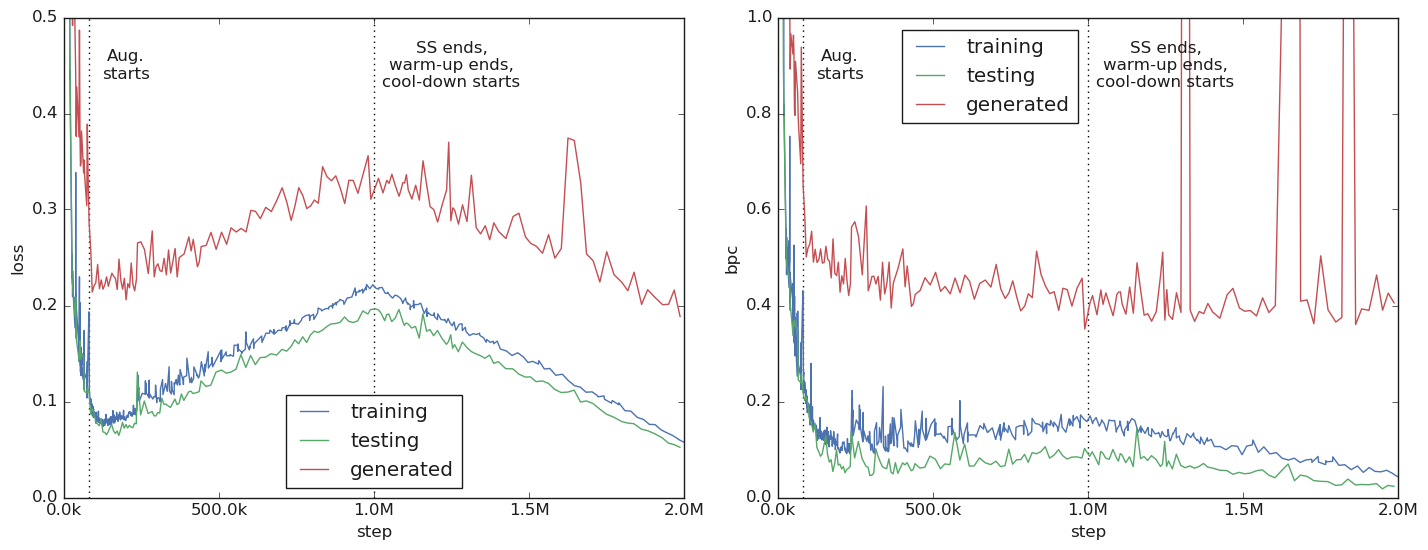}
\end{figure}

Teacher forcing for the strings remains the best training scheme, but lack of the RNN-based \texttt{Tuple} decoder significantly cripples the model's capacity of capturing correlations. Augmented training manages to improve over the low baseline, but multiscale VAE fails to yield improvements. Most likely, the pass-through model lacks the capacity to learn what zip code is associated with given coordinates (or vice versa) and resorts to encode both zip code and coordinates in the latent vector in parallel. Generated loss/BPC increases during most of the training process for these pass-through models (Fig~\ref{fig:simple_teacher_generated_loss}), unless augmented training is employed (Fig~\ref{fig:simple_augmented_generated_loss}).

\clearpage

\section{AAE (Adversarial Autoencoder) model}
\label{appendix:adversarial_autoencoder_model}

We have tested using the Adversarial Autoencoder (AAE) framework \citep{DBLP:journals/corr/MakhzaniSJG15} with the tree recursive model for this problem. To keep the number of parameters approximately the same, we repurposed the standard deviation network described in Sec~\ref{sec:standard_deviation_network} for the role of discriminator by replacing the final sigmoid layer with a $128 \times 2$ softmax layer to distinguish the latent vector generated by the deterministic encoder from the random vector drawn from the unit Gaussian distribution. We found that despite running the same Tuple SS + String TF training scheme, an AAE model failed to capture the zip-coordinate correlations (p-value stats turned out to be mean = 0.0576, median = 0, standard deviation = 0.183 after 2M autoencoder training steps and 2M discriminator training steps on the same learning rate schedule as the VAE models. Cross-entropy adversarial loss was given the same weight as the reconstruction loss). %

\section{WAE (Wasserstein Autoencoder) model}
\label{appendix:wasserstein_autoencoder_model}

We have also tested using the Wasserstein Autoencoder \citep{2017arXiv171101558T} with the tree recursive model for this problem. In our implementation, we make the simplifying assumption that the latent vectors follow a multivariate Gaussian distribution and proceed to estimate its mean vector $m$ and covariance matrix $\Sigma$. The $W_2$ Wasserstein distance squared from this multivariate Gaussian distribution to the 128-dim unit Gaussian distribution is then given by 

$$W_2(\mathcal{N}(m,\Sigma);\mathcal{N}(0,I))^2 =\Vert m \Vert_2^2 +\Vert\Sigma^{1/2}-I\Vert_{Frobenius}^2$$

where $\Vert \Vert_2^2$ is the $L_2$ norm, $I$, is the identity matrix, and $\Vert \Vert_{Frobenius}^2$ is the matrix Frobenius norm. We also found that a WAE model running the same Tuple SS + String TF training scheme failed to capture the zip-coordinate correlations (p-value stats turned out to be mean = 0.0240, median = 0, standard deviation = 0.112 after 2M training steps on the same learning rate schedule as the VAE models. The $W_2^2$ Wasserstein distance squared latent loss was given the weight 0.128 per latent dimension). As pointed out by \citep{2017arXiv171101558T}, AAE can be considered a special case of WAE, so we find these results to be consistent. We suspect that this problem requires the framework to ensure the continuity of the decoder w.r.t. individual training examples, rendering generative model frameworks that only regualrize the latent vector distribution insufficient. %

\section{Alternative Multiscale VAE formulations}
\label{appendix:alternative_multiscale}

In Sec~\ref{sec:multiscale}, we initialize $n_{\mathrm{KL\_weight}}$ to be equally spaced on the interval $(0, \beta_{\mathrm{max}}]$ for simplicity, and we find the optimized multiscale VAE's behavior to be sensible. That is, the model accurately captures the zip-coordinate correlations and considers street name beyond its capacities to encode as it varies within the reconstruction loss of the coordinates. We also note that adding augmented training partially restores street name autoencoding by making up its own details, with a price in terms of correlation accuracy and street name realism. However, we are still curious whether multiscale VAE can be forced to encode more details such as street names by setting more workers to train with lower $\beta$ values.

Inspired by the exponential decay term $(1 - p_{\mathrm{sampled}})$ in the augmented objective function Eq \eqref{eq:augmented}, the most obvious alternative is to use geometric spacing instead of linear spacing. For example, with common ratio $r = 0.9$ and $n_{\mathrm{KL\_weight}} = 32$, we have $\beta_0 = 0.9^{31} \beta_{\mathrm{max}}, \beta_1 = 0.9^{30} \beta_{\mathrm{max}}, \dots, \beta_{30} = 0.9 \beta_{\mathrm{max}}, \beta_{31} = \beta_{\mathrm{max}}$ for 32 workers. Again comparing the multiscale objective function Eq \eqref{eq:multiscale} with the augmented objective function Eq \eqref{eq:augmented}, we noticed that although the KL-divergence weight $\beta_i$ is usually applied to the KL-divergence directly, the multiscale objective function with geometric spacing would match the augmented objective function Eq \eqref{eq:augmented} more closely if we invert the weights and divide the reconstruction loss by $\beta_i$ instead:

\begin{equation}
\label{eq:multiscale_inverted}
\begin{aligned}
\sum_{i=0}^{\mathrm{kl\_weight\_levels} - 1} \frac{1}{\beta_i} \mathbb{E}[\log p_\theta(x|z^{(i)})] &- KL(q_{\lambda, i}(z^{(i)}|x)||p(z^{(i)})) \\
\text{where } z^{(i)} &\sim \mathcal{N}(\mu_\lambda(x), \sigma_{\lambda, i}(\mu_\lambda(x)))
\end{aligned}
\end{equation}

For a plain $\beta$-VAE this does not matter, especially when a gradient-normalizing optimizer like Adam is used. For a multiscale VAE however, the difference in formulation changes how gradients from workers training at different scales are weighted. For the result below, we tested both. Training scheme is fixed to be Tuple SS + String TF, and we measure both the zip-coordinate p-value stats and the average Levenshtein distance per character $\bar{d}_{\mathrm{Levenshtein}}$ between the original street name and its reconstruction.

\begin{table}[ht]
\centering
  \caption{Multiscale VAE with geometric spacing (weight inverted)}\label{tab:multiscale_geo_inverted}
  \begin{adjustbox}{center}
  \begin{tabular}{| c | c | c | c | c | c | c |}
    \hline
    $r$ & $\beta_{\mathrm{max, start}}$ & $\beta_{\mathrm{max, end}}$ & mean & median & stddev & $\bar{d}_{\mathrm{Levenshtein}}$ \\ %
    \hline
    \multirow{3}{*}{0.9} & 1.28 & 0.64 & 0.425 -- \textbf{0.445} & 0.403 -- \textbf{0.448} & 0.362 -- 0.368 & 0.601 -- 0.732 \\ %
    \cline{2-7}
    & 2.56 & 1.28 & 0.441 -- 0.488 & 0.433 -- 0.538 & 0.376 -- 0.379 & 0.904 -- 0.918 \\ %
    \cline{2-7}
    & 5.12 & 2.56 & 0.395 -- 0.395 & 0.295 -- 0.314 & 0.370 -- 0.364 & 0.922 -- 0.917 \\ %
    \hline
    \multirow{3}{*}{0.8} & 1.28 & 0.64 & 0.264 -- 0.264 & 0.0830 -- 0.0939 & 0.322 -- 0.317 & 0.0354 -- 0.0475 \\ %
    \cline{2-7}
    & 2.56 & 1.28 & 0.308 -- 0.363 & 0.184 -- 0.291 & 0.325 -- 0.334 & 0.264 -- 0.391 \\ %
    \cline{2-7}
    & 5.12 & 2.56 & 0.400 -- 0.403 & 0.354 -- 0.360 & 0.362 -- 0.358 & 0.594 -- 0.710 \\ %
    \hline
  \end{tabular}
  \end{adjustbox}
\end{table}

\begin{table}[ht]
\centering
  \caption{Multiscale VAE with geometric spacing}\label{tab:multiscale_geo}
  \begin{adjustbox}{center}
  \begin{tabular}{| c | c | c | c | c | c | c |}
    \hline
    $r$ & $\beta_{\mathrm{max, start}}$ & $\beta_{\mathrm{max, end}}$ & mean & median & stddev & $\bar{d}_{\mathrm{Levenshtein}}$ \\ %
    \hline
    \multirow{3}{*}{0.9} & 1.28 & 0.64 & 0.342 -- 0.461 & 0.213 -- 0.477 & 0.353 -- 0.351 & 0.479 -- 0.725 \\ %
    \cline{2-7}
    & 2.56 & 1.28 & 0.493 -- 0.494 & 0.548 -- 0.549 & 0.385 -- 0.380 & 0.910 -- 0.921 \\ %
    \cline{2-7}
    & 5.12 & 2.56 & 0.364 -- 0.460 & 0.251 -- 0.471 & 0.359 -- 0.376 & 0.920 -- 0.920 \\ %
    \hline
    \multirow{3}{*}{0.8} & 1.28 & 0.64 & 0.247 -- 0.269 & 0.0563 -- 0.0998 & 0.316 -- 0.320 & 0.0620 -- 0.0714  \\ %
    \cline{2-7}
    & 2.56 & 1.28 & 0.325 -- 0.332 & 0.212 -- 0.223 & 0.332 -- 0.336 & 0.315 -- 0.278 \\ %
    \cline{2-7}
    & 5.12 & 2.56 & 0.384 -- 0.403 & 0.313 -- 0.348 & 0.357 -- 0.371 & 0.565 -- 0.855 \\ %
    \hline
  \end{tabular}
  \end{adjustbox}
\end{table}

The following is the street name reconstructions for the first 10 training examples by the run marked by the bold font:

\begin{verbatim}
"HARTS RD" -> "HARTS RD"
"SECOND ST" -> "SPRING ST"
"LIME KILN RD" -> "LAKE DAVIS RD"
"JOHNSON HILL RD" -> "BAY HILL RD"
"JACKSON CROSS RD" -> "MOUNTAINSON AVE"
"WAUGH FARM RD" -> "RUNNS LN"
"PAINT WORKS RD" -> "PORTECHAN WINOORD RD"
"FURLONG RD" -> "FLATC ST"
"SABIN ST" -> "SOUTHWIN TER"
"POISSON DR" -> "POPE PKWY"
\end{verbatim}

We can see that multiscale VAE can indeed be tuned to encode more details such as street names. However, there remains a trade-off between accurate zip-coordinate correlations and details such as street names, and fine-tuned multiscale VAE is no better than multiscale VAE trained with augmented training (Appendix \ref{appendix:multiscale_augmented}) at comparable correlation accuracy. Interestingly, weight inversion does seem to make the model more reliable at street name reconstruction with comparable zip-coordinate correlation accuracy.

A more radical alternative is to assign a different target KL-divergence value $C$ for each worker a la \citep{burgess2018understanding}, instead of a different KL-divergence weight that still allows each worker to find its own trade-off between latent loss and reconstruction loss. We were surprised to find out, however, that it does not work to specify the target total KL-divergence in terms of L1 loss $\gamma |KL(q_\lambda(z|x)||p(z)) - C|$. Instead, we only punish the model for going over the capacity budget, and trust the reconstruction loss alone to use as much capacity budget as possible with loss term $\gamma \max(KL(q_\lambda(z|x)||p(z)) - C, 0)$, so we have the following objective function:

\begin{equation}
\label{eq:multiscale_capacity_penalty}
\begin{aligned}
\sum_{i=0}^{\mathrm{kl\_weight\_levels} - 1} \mathbb{E}[\log p_\theta(x|z^{(i)})] &- \gamma \max(KL(q_\lambda(z|x)||p(z)) - C_i, 0) \\
\text{where } z^{(i)} &\sim \mathcal{N}(\mu_\lambda(x), \sigma_{\lambda, i}(\mu_\lambda(x)))
\end{aligned}
\end{equation}

For the result below, capacity penalty weight $\gamma$ is set to be 128 per latent dimension while the reconstruction loss is still the weighted average of nat-per-character and mean squared error loss terms to make sure that capacity budget is respected. The target capacities are specified with minimum capacity $C_{min}$ and capacity increment $C_{increment}$. For example, if $C_{min} = 10$ and $C_{increment} = 0.5$, the 32 workers run with capacity budgets $C_0 = 10, C_1 = 10.5, C_2 = 11, \dots, C_{31} = 25.5$. Target capacities remain fixed throughout the training process for these runs.

\begin{table}[ht]
\centering
  \caption{Multiscale VAE with target capacities}\label{tab:multiscale_capacity}
  \begin{tabular}{| c | c | c | c | c | c |}
    \hline
    $C_{min}$ & $C_{increment}$ & mean & median & stddev & $\bar{d}_{\mathrm{Levenshtein}}$ \\ %
    \hline
    \multirow{4}{*}{10} & 0.2 & 0.434 -- 0.449 & 0.402 -- 0.434 & 0.382 -- 0.375 & 0.913 -- 0.914 \\ %
    \cline{2-6}
    & 0.35 & 0.403 -- 0.414 & 0.339 -- 0.372 & 0.364 -- 0.359 & 0.734 -- 0.759 \\ %
    \cline{2-6}
    & 0.5 & 0.276 -- 0.301 & 0.0874 -- 0.125 & 0.332 -- 0.341 & 0.483 -- 0.517 \\ %
    \cline{2-6}
    & 1.0 & 0.216 -- 0.223 & 0.0265 -- 0.0388 & 0.301 -- 0.302 &  0.118 -- 0.125 \\ %
    \hline
    15 & 0.2 & 0.319 -- 0.343 & 0.153 -- 0.223 & 0.350 -- 0.347 & 0.642 -- 0.628 \\ %
    \hline
  \end{tabular}
\end{table}

We observe the same trade-off between accurate zip-coordinate correlations and details such as street names and are not able to get better result. Perhaps it is harder to tune multiscale VAE with target capacities since the model is not allowed to make trade-offs between reconstruction loss and latent loss on its own.

\section{Multiscale VAE + augmented training experiments}
\label{appendix:multiscale_augmented}

We explore whether multiscale VAE can be improved by augmented training. The experiments below are based on the hyperparameters optimized in Sec~\ref{sec:multiscale} ($\beta_{\mathrm{max, start}} = 1.28, \beta_{\mathrm{max, end}} = 0.64$). To compensate slow data generation, we always shut down the training process at \texttt{gen\_start\_step} and bring it back up with $n_{\mathrm{augmented}}$ = 256 and 512 workers. Due to the observed variability, we always run experiments with the same hyperparameters twice.

\begin{table}[ht]
  \centering
  \caption{Multiscale VAE + augmented training performance}\label{tab:multiscale_augmented}
  \begin{tabular}{| c | c | c | c | c | c |}
    \hline
    \texttt{gen\_start\_step} & $p_{\mathrm{sampled}}$ & mean & median & stddev & $\bar{d}_{\mathrm{Levenshtein}}$ \\ %
    \hline
    \multicolumn{6}{|c|}{Tuple SS + String TF} \\
    \hline
    $10^6$ & \multirow{2}{*}{$\nicefrac{1}{5}$} & 0.465 -- \textbf{0.471} & 0.471 -- \textbf{0.491} & 0.377 -- 0.383 & 0.770 -- 0.780 \\ %
    \cline{1-1} \cline{3-6}
    $2 \times 10^5$ & & 0.444 -- 0.457 & 0.425 -- 0.466 & 0.385 -- 0.383 & 0.719 -- 0.662 \\ %
    \hline
    \multicolumn{6}{|c|}{Scheduled Sampling} \\
    \hline
    \multirow{3}{*}{$8 \times 10^5$} & $\nicefrac{1}{5}$ & 0.358 -- 0.385 & 0.196 -- 0.267 & 0.376 -- 0.381 & 0.427 -- 0.441 \\ %
    \cline{2-6}
    & $\nicefrac{1}{3}$ & 0.338 -- 0.378 & 0.154 -- 0.270 & 0.369 -- 0.377 & 0.423 -- 0.407 \\ %
    \cline{2-6}
    & $\nicefrac{1}{2}$ & 0.357 -- 0.382 & 0.217 -- 0.268 & 0.368 -- 0.377 & 0.378 -- 0.373 \\ %
    \hline
  \end{tabular}
\end{table}

\begin{figure}[ht]
\centering
\caption{Loss (left) and BPC (right) for training data, testing data, and generated samples during the training process of multiscale VAE + augmented training on the worker with the lowest $\beta$ value $\beta_0 = \frac{1}{32}\beta_{\mathrm{max}}$.}\label{fig:multiscale_augmented_generated_loss} %
\includegraphics[width=\textwidth]{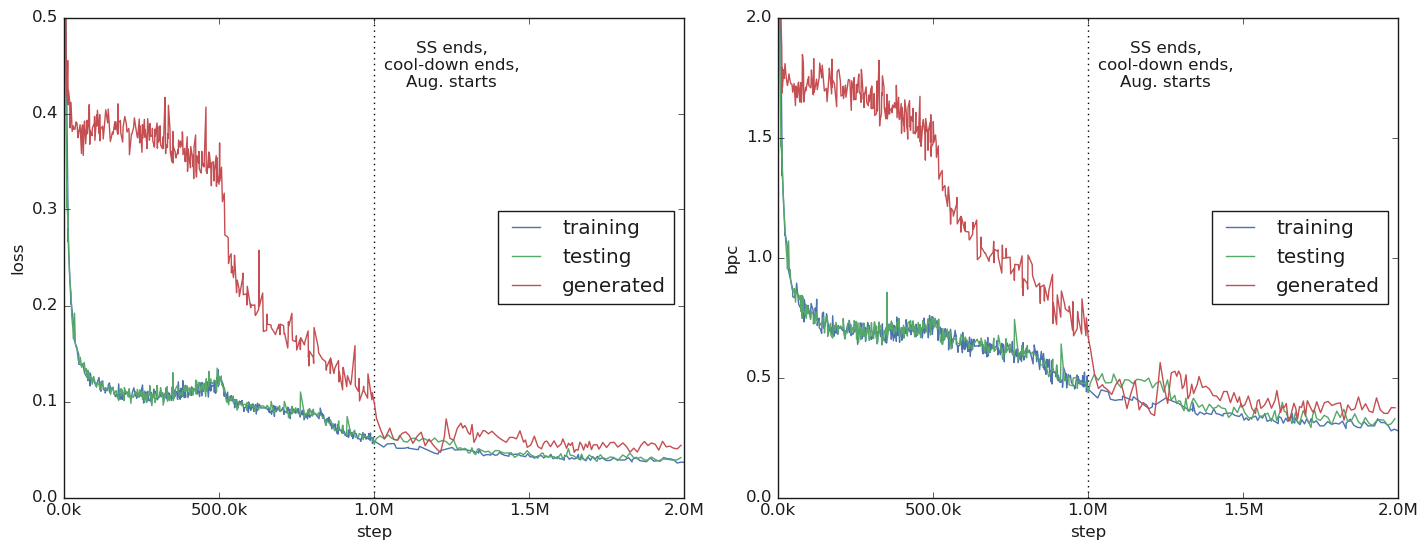}
\end{figure}

Multiscale VAE alone is capable of capturing the zip-coordinate correlations and generated loss decreases during the training process under the Tuple SS + String TF training scheme, so it turns out to be beneficial to delay the start of augmented training until at least 1M steps. Optimized multiscale VAE + augmented training features a good compromise between the two with accurate zip-coordinate correlations, low generated loss (Fig~\ref{fig:multiscale_augmented_generated_loss}, taken from the run of Table~\ref{tab:multiscale_augmented} marked by the bold font), and partially restored street name reconstructions. We noticed that even though optimized multiscale VAE + augmented training usually reconstructs the first few letters of the street name, sometimes the corresponding embedding (mean) vector actually encodes a different but typical street name, especially when the training example features an unusual street name like "LIME KILN RD" or "WAUGH FARM RD". We believe this is another manifestation of the additive gravitational pull of the training examples, in combination with augmented training..

\begin{figure}[ht]
\centering
\caption{Box plot of p-values over repeated encoding and decoding (multiscale VAE + augmented training).}\label{fig:multiscale_augmented_repeated_boxes}
\includegraphics[width=\textwidth]{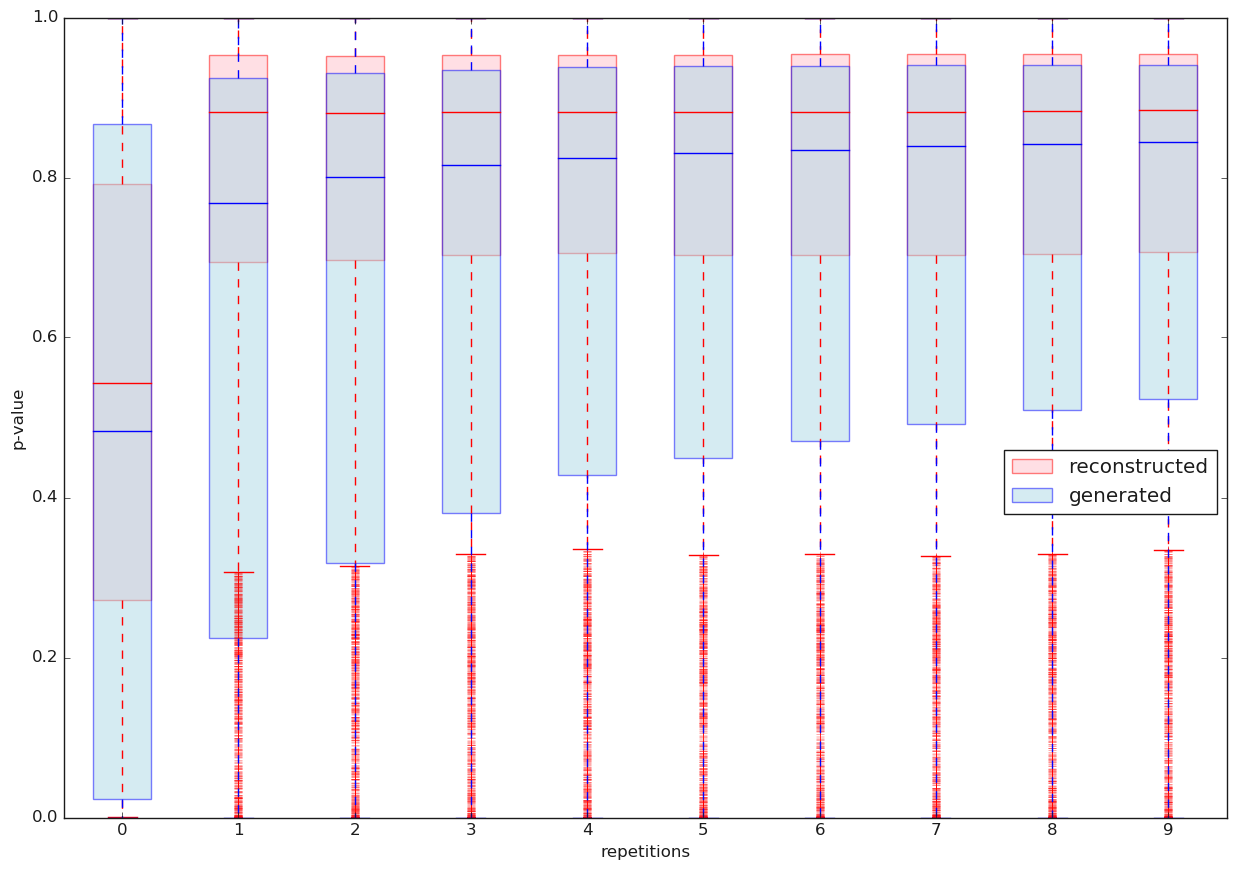}
\end{figure}

\begin{figure}[ht]
\centering
\caption{Number of generated street names present in the training data out of 10000 samples of $\tilde{x}^{(n)}$ over repeated encoding and decoding (multiscale VAE only + augmented training). The proportion (51\%) is higher than that of optimized $\beta$-VAE (44\%, Fig~\ref{fig:warm_up_teacher_str_street_names}) and stays roughly constant throughout repeated encoding/decoding.}\label{fig:multiscale_augmented_teacher_str_street_names}
\includegraphics[width=\textwidth]{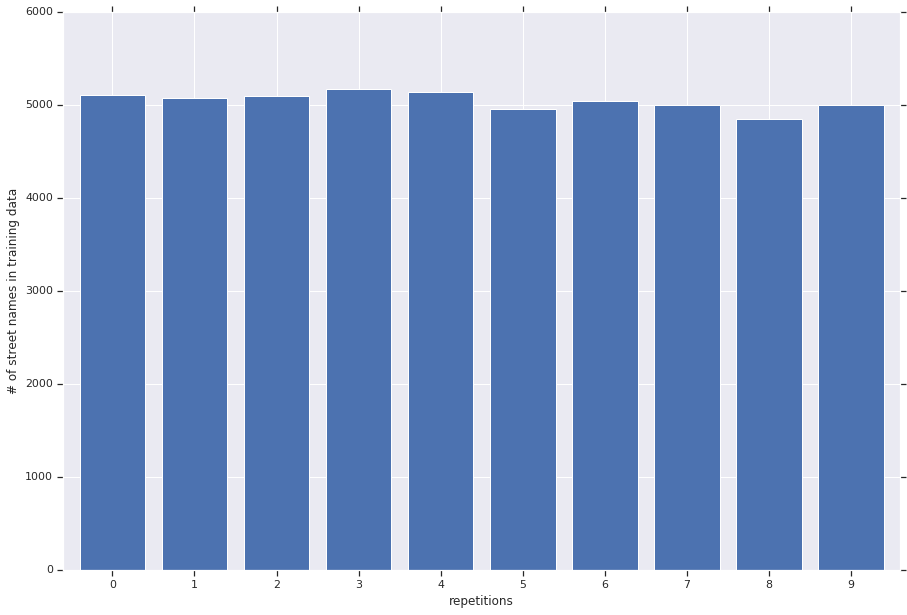}
\end{figure}

\begin{figure}[ht]
\centering
\caption{Interpolation between the first 2 training examples by the multiscale VAE with augmented training.}\label{fig:multiscale_augmented_interpolation}
\includegraphics[width=\textwidth]{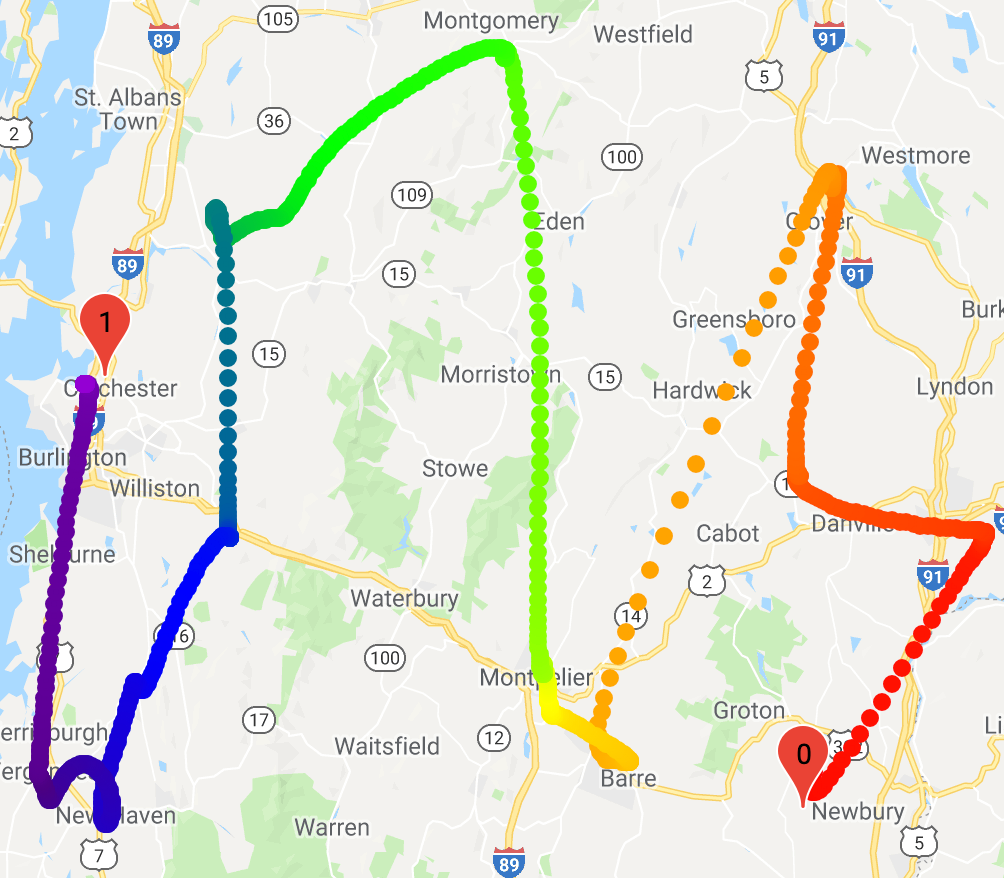}
\end{figure}

\clearpage

With the speculation that sampled latent vector loses information about the training example faster when $\beta$ is higher, the experiments below have adaptive $p_{\mathrm{sampled}}$ on the interval $(p_{\mathrm{sampled, min}}, p_{\mathrm{sampled, max}}]$, equally spaced on the linear scale. For example, if $p_{\mathrm{sampled, min}} = 0.2, p_{\mathrm{sampled, max}} = 1$, and $n_{\mathrm{KL\_weight}} = 32$, values of $p_{\mathrm{sampled}}$ form an arithmetic sequence such that workers run with parameters $(\beta, p_{\mathrm{sampled}}) = (\frac{1}{32}\beta_{\mathrm{max}}, 0.225), (\frac{2}{32}\beta_{\mathrm{max}}, 0.25), \dots, (\beta_{\mathrm{max}}, 1)$.

\begin{table}[ht]
  \caption{Multiscale VAE + augmented training performance (linearly-spaced)}
  \begin{adjustbox}{center}
  \begin{tabular}{| c | c | c | c | c | c | c |}
    \hline
    \texttt{gen\_start\_step} & $p_{\mathrm{sampled, min}}$ & $p_{\mathrm{sampled, max}}$ & mean & median & stddev & $\bar{d}_{\mathrm{Levenshtein}}$ \\ %
    \hline
    \multicolumn{7}{|c|}{Tuple SS + String TF} \\
    \hline
    $10^6$ & \multirow{2}{*}{$\nicefrac{1}{5}$} & \multirow{2}{*}{1} & 0.428 -- 0.484 & 0.420 -- 0.528 & 0.357 -- 0.376 & 0.621 -- 0.715 \\ %
    \cline{1-1} \cline{4-7}
    $2 \times 10^5$ & & & 0.350 -- 0.408 & 0.233 -- 0.346 & 0.355 -- 0.374 & 0.486 -- 0.507 \\ %
    \hline
    \multicolumn{7}{|c|}{Scheduled Sampling} \\
    \hline
    \multirow{5}{*}{$8 \times 10^5$} & $\nicefrac{1}{10}$ & \multirow{5}{*}{1} & 0.374 -- 0.393 & 0.238 -- 0.316 & 0.380 -- 0.376 & 0.527 -- 0.426 \\ %
    \cline{2-2} \cline{4-7}
    & $\nicefrac{1}{8}$ & & 0.381 -- 0.390 & 0.262 -- 0.283 & 0.381 -- 0.381 & 0.446 -- 0.448 \\ %
    \cline{2-2} \cline{4-7}
    & $\nicefrac{1}{5}$ & & 0.382 -- 0.417 & 0.279 -- 0.357 & 0.375 -- 0.386 & 0.392 -- 0.536 \\ %
    \cline{2-2} \cline{4-7}
    & $\nicefrac{1}{3}$ & & 0.358 -- 0.391 & 0.218 -- 0.291 & 0.368 -- 0.380 & 0.391 -- 0.568 \\ %
    \cline{2-2} \cline{4-7}
    & $\nicefrac{1}{2}$ & & 0.361 -- 0.365 & 0.226 -- 0.246 & 0.370 -- 0.368 & 0.441 -- 0.369 \\ %
    \hline
  \end{tabular}
  \end{adjustbox}
\end{table}

We have also tested multiscale VAE + augmented training (scheduled sampling) with less $n_{\mathrm{augmented}}$ and adaptive $p_{\mathrm{sampled}}$ on the interval $(p_{\mathrm{sampled, min}}, p_{\mathrm{sampled, max}}]$ equally spaced on the log scale. For example, if $p_{\mathrm{sampled, min}} = 0.2, p_{\mathrm{sampled, max}} = 1$, and $n_{\mathrm{KL\_weight}} = 32$, values of $p_{\mathrm{sampled}}$ form a geometric sequence such that workers run with parameters $(\beta, p_{\mathrm{sampled}}) = (\frac{1}{32}\beta_{\mathrm{max}}, 5^{-\frac{31}{32}}), (\frac{2}{32}\beta_{\mathrm{max}}, 5^{-\frac{30}{32}}), \dots, (\beta_{\mathrm{max}}, 1)$. They do not work better and turned out to be irrelevant in the context of later findings, so we report them here for  completeness. Just like previous experiments, we always shut down and bring the training process back up at \texttt{gen\_start\_step} = $8 \times 10^5$ with $n_{\mathrm{augmented}} \times 2$ workers.

\begin{table}[ht]
  \caption{Multiscale VAE + augmented training performance (geometrically-spaced)}
  \begin{adjustbox}{center}
  \begin{tabular}{| c | c | c | c | c | c | c |}
    \hline
    $n_{\mathrm{augmented}}$ & $p_{\mathrm{sampled, min}}$ & $p_{\mathrm{sampled, max}}$ & mean & median & stddev & $\bar{d}_{\mathrm{Levenshtein}}$ \\ %
    \hline \hline
    \multicolumn{7}{|c|}{Scheduled Sampling} \\
    \hline
    16 & \multicolumn{2}{|c|}{\multirow{5}{*}{1}} & 0.373 -- 0.380 & 0.251 -- 0.283 & 0.374 -- 0.368 & 0.542 -- 0.435 \\ %
    \cline{1-1} \cline{4-7}
    32 & \multicolumn{2}{|c|}{} & 0.373 -- 0.383 & 0.257 -- 0.303 & 0.371 -- 0.366 & 0.529 -- 0.410 \\ %
    \cline{1-1} \cline{4-7}
    64 & \multicolumn{2}{|c|}{} & 0.367 -- 0.375 & 0.246 -- 0.269 & 0.369 -- 0.370 & 0.472 -- 0.571 \\ %
    \cline{1-1} \cline{4-7}
    128 & \multicolumn{2}{|c|}{} & 0.368 -- 0.378 & 0.246 -- 0.273 & 0.370 -- 0.372 & 0.445 -- 0.524 \\ %
    \cline{1-1} \cline{4-7}
    \multirow{7}{*}{256} & \multicolumn{2}{|c|}{} & 0.318 -- 0.388 & 0.175 -- 0.295 & 0.340 -- 0.377 & 0.575 -- 0.422 \\ %
    \cline{2-7}
    & \multicolumn{6}{|c|}{Geometric sequence} \\
    \cline{2-7}
    & $\nicefrac{1}{16}$ & \multirow{5}{*}{1} & 0.374 -- 0.379 & 0.228 -- 0.267 & 0.382 -- 0.375 & 0.427 -- 0.583 \\ %
    \cline{2-2} \cline{4-7}
    & $\nicefrac{1}{10}$ & & 0.361 -- \textbf{0.405} & 0.224 -- \textbf{0.321} & 0.369 -- 0.386 & 0.573 -- 0.631 \\ %
    \cline{2-2} \cline{4-7}
    & $\nicefrac{1}{8}$ & & 0.358 -- 0.389 & 0.249 -- 0.286 & 0.359 -- 0.378 & 0.582 -- 0.519 \\ %
    \cline{2-2} \cline{4-7}
    & $\nicefrac{1}{5}$ & & 0.371 -- 0.389 & 0.268 -- 0.276 & 0.367 -- 0.381 & 0.397 -- 0.537 \\ %
    \cline{2-2} \cline{4-7}
    & $\nicefrac{1}{2}$ & & 0.361 -- 0.377 & 0.232 -- 0.266 & 0.368 -- 0.369 & 0.628 -- 0.432 \\ %
    \hline
  \end{tabular}
  \end{adjustbox}
\end{table}

\clearpage

\section{Other negative results}

\begin{enumerate}
  \item It may be counter-intuitive to initialize character embedding with uniform distribution between 0 and 1 and use the sigmoid function as the activation function for the \texttt{ScalarTuple} encoder, while CELU with $\alpha=3$ is used for most of the model. However, changing them to cover more of the unit Gaussian distribution by their respective embeddings doesn't yield any improvement.
  \item It's critical to bias the information diffusion in the latent space towards spreading information from the training examples for augmented training. Generation quality is improved only when augmented latent vectors are initialized from the sampled latent vectors of the training examples.
  \item Replacing $p_{\mathrm{sampled}}$ with $p_{fake}(x)$ given by a discriminator for augmented training doesn't work as intended. Without a way to use gradient descent to increasingly confuse the discriminator, what does confuse the discriminator tends to be near-exact copies of training examples. Variants that keep a fraction of the augmented latent vectors that correspond to lower $p_{fake}(x)$ don't work either. %
  \item Since we maintain full data parallelism for multiscale VAE training, a training batch on a worker is always used with the same $\beta$ value and standard deviation network. Using a training batch with multiple $\beta$ values and standard deviation networks, either deterministically or randomly, doesn't yield further improvement within the same training budget. %
  \item Early attempts to replace multiple standard deviation networks $\sigma_i(\mu)$ with one parameterized by the $\beta$ value $\sigma(\mu, \beta)$, e.g. making one or more of its fully-connected layers $129 \times 128$ and appending $\frac{\beta}{\beta_{\mathrm{max}}}$ to their input, do not work. It may be inherently difficult to learn a shared representation across multiple scales. %
\end{enumerate}

\end{document}